\documentclass[a4paper,10pt,twocolumn]{article}

\usepackage{amsmath}
\usepackage{amssymb}
\usepackage[dvipdfmx]{graphicx}
\usepackage{ifthen}
\usepackage{cite}
\usepackage{verbatim}
\usepackage[usenames]{color}
\usepackage{slashbox}
\usepackage{mdwlist}
\usepackage{url}

\begin{document}

\title{
	Markov Chain Monte Carlo for Arrangement of Hyperplanes\\
	in Locality-Sensitive Hashing.
}
\setcounter{footnote}{0}
\author{
Yui~Noma
\footnotemark[1]
\and
Makiko~Konoshima\footnote{Software Systems Laboratories, FUJITSU LABORATORIES LTD.  1-1,
Kamikodanaka 4-chome, Nakahara-ku Kawasaki, 211-8588 Japan.}
\footnote{E-mail: \texttt{makiko@jp.fujitsu.com}}
}
\date{}
\maketitle

\begin{abstract}


Since Hamming distances can be calculated by bitwise computations,
 they can be calculated with less computational load than L2 distances.
Similarity searches can therefore be performed faster in Hamming distance space.
The elements of Hamming distance space are bit strings.
On the other hand,
 the arrangement of hyperplanes induce the transformation from the feature vectors
 into feature bit strings.
This transformation method is a type of locality-sensitive hashing that has been
 attracting attention as a way of performing approximate similarity searches at high speed.
Supervised learning of hyperplane arrangements allows us to obtain a method
 that transforms them into feature bit strings reflecting the information of
 labels applied to higher-dimensional feature vectors.
In this paper, we propose a supervised learning method for hyperplane arrangements
 in feature space that uses a Markov chain Monte Carlo (MCMC) method.


We consider the probability density functions used during learning,
 and evaluate their performance.
We also consider the sampling method for learning data pairs needed in learning,
 and we evaluate its performance.
We confirm that the accuracy of this learning method when using a suitable probability
 density function and sampling method is greater than the accuracy of existing learning methods.

{\bf Keyword:}
 Higher dimensional feature vector,
 Locality-sensitive hashing,
 Arrangement of hyperplanes,
 Similarity search,
 Markov chain Monte Carlo,
 Low-temperature limit
\end{abstract}

\section{Introduction}



Unstructured data such as audio and images includes complex content.
This makes it difficult to search for unstructured data directly.
A common approach has therefore been to perform searches based on
 feature vectors extracted from unstructured data.
To reflect the complexity of unstructured data,
 these feature vectors generally consist of higher-dimensional data
 with hundreds or even thousands of dimensions.


There are a wide range of applications for high-speed similarity searching
 using higher-dimensional feature quantities extracted from unstructured data.
Examples include authentication of people by fingerprint recognition,
 speech recognition in call centers, management of products and components based on
 CAD data, and detecting abnormal situations from surveillance video.
For these applications, there are two things that are very important.
One is a high-speed similarity search method.
The other is a data structure that permits high-speed similarity searching,
 and a method for extracting feature quantities that reflect the properties of unstructured data.


The high-speed similarity search method is described first.
To perform a similarity search, the feature space should be a metric space.
In most cases, the feature space is treated as an L2 metric space.
Many studies have devised an index structure aimed at performing similarity searches
 at high speed.
For example, the literatures~\cite{KDTree, iDistance} are two of them.
However, in higher-dimensional space, due to the so-called 
``the curse of dimensionality'', all distances between data items are of similar size.
Consequently, searches in higher-dimensional data using these methods end up
 having processing times that are similar to those of searches performed without
 using a special index~\cite{Weber1998DimensionCurse}.


Hamming distances can be calculated by bitwise operations,
 which means that similarity searches are fast in Hamming metric space
 without using a specific index structure.
In a method called locality-sensitive hashing~\cite{LSH_IndykMotwani},
 the feature vectors are transformed into bit strings.
For this transformation, methods that involve the use of hyperplanes
 in feature space have been intensively studied~\cite{LSH_RandomProjection,LSH_FS1,LSH_FS2,PCAH}.
In these methods, multiple hyperplanes are considered as a means of partitioning the feature space.
A bit string is assigned to each partitioned region,
 determined from the orientations of hyperplanes.
Feature vectors extracted from the data are allocated in the same way as bit strings assigned
 to the regions that include the feature vectors.
Similar feature vectors are included in neighboring regions,
 so the bit strings allocated to these feature vectors are similar and
 are separated by small Hamming distances.
In the following, we will use the term ``hashing'' to refer to the process of transforming
 higher-dimensional feature vectors into feature bit strings.


Next, we consider a data structure that can be searched at high speed,
 and a method for extracting feature quantities reflecting the properties of unstructured data.
From the above discussion, we decided to use bit strings as feature quantities,
 since these are data structures that can be searched at high speed.
When data has been labeled, supervised learning can be used to extract feature bit strings
 that reflect the labeled  information.
In the following, feature quantities that reflect the labeled information are described
 as high-precision quantities.
Also, a learning method that can extract high-precision feature quantities is described
 as a high-performance learning method.


Studies aimed at increasing the precision of feature quantities associated
 with the hyperplane hashing method include
 the following references~\cite{MLH,SIMBA_KN,LIFT_KN}.
In learning, the normal vectors of the hyperplanes are determined by making
 the Hamming distances smaller between data pairs with a common label and larger
 between data pairs that do not have any common label.
As the number of bits increases, the degree of freedom also increases
 so that greater precision becomes possible.


Based on this reasoning, we can draw the following conclusions regarding
 high-precision similarity searching of large quantities of unstructured data.
High-speed similarity searches are achieved by using bit strings in Hamming metric space
 as feature quantities.
High precision is achieved by using a large number of bits and performing
 supervised learning with labeled data as the training data.
However, to reflect the complexity of unstructured data,
 it should be noted that a single item of unstructured data will not necessarily have just one label.


In this paper, apart from the use of feature bit strings,
 no consideration is given to the processing time of the similarity search.
Our main focus is on using supervised learning to improve the precision of feature bit strings.


The method proposed in this paper performs supervised learning using MCMC.
The transformation of feature vectors into feature bit strings is a discontinuous mapping.
This makes it impossible to perform na\"{i}ve learning based on gradients.
Another approach involves introducing a loss function so that the transformation method
 can be approximated by a continuous function.
However, the only loss functions found so far are strongly dependent on the properties of the data set.
In our proposed method, each normal vector is regarded as a particle on a unit sphere in feature space,
 and a random walk is performed on this unit sphere.
In the random walk, a discontinuous function can be treated as an evaluation function.
We also considered sampling methods for training data pairs and evaluation functions for use in learning.


This paper is structured as follows.
First, in section~\ref{sec_relatedWork} we describe the existing learning methods.
Then in section~\ref{sec_proposal}, we describe our proposed method.
We considered evaluation functions needed during learning, and sampling methods for training data pairs.
In section~\ref{sec_experimentsAndResult}, we perform experiments using various data sets.
At the same time, we also evaluate the evaluation functions and the sampling methods.
In section~\ref{sec_kizontonosa}, we show that the proposed learning method performs
 better than existing methods.
Finally, in section~\ref{sec_conclusion} we summarize our work and discuss the future prospects of this approach.

\section{Background and related work}
\label{sec_relatedWork}


In this section, we describe the use of hyperplanes for locality-sensitive hashing,
 which is the basis of the proposed technique.
We then describe some related existing techniques.

\subsection{Conventional locality-sensitive hashing with hyperplanes}


The hashing method using hyperplanes is described below.
A space $V$ in which there are higher-dimensional feature quantities is regarded
 as an $N$-dimensional vector space.
The configurations of multiple hyperplanes in $V$ are referred to as hyperplane arrangements.


Consider $B$ hyperplanes passing through the origin of $V$.
A hyperplane passing through the origin is identified by its normal vector.
An $N$-dimensional feature vector $\vec{x}$ is transformed into a bit string by registering
 a 1 if its dot product with each normal vector is positive,
 and a zero otherwise.
Therefore, the length of the bit string is equal to the number of hyperplanes $B$.


A hyperplane that does not pass through the origin can easily be constructed from a hyperplane that does.
In reference~\cite{LIFT_KN}, an experiment is performed where the hashing of
 hyperplanes that do not pass through the origin is learned by learning
 the hashing of hyperplanes that do pass through the origin.
When developing a new learning method for hyperplanes,
 it is easier to work with hyperplanes that pass through the origin.
In the following discussion, therefore,
 all hyperplanes are assumed to pass through the origin.


When labels have been applied to the data,
 it is sometimes the case that the angles or L2 distances do not exhibit
 a suitable degree of dissimilarity.
In such cases, the hyperplanes can be determined by supervised learning. 
In supervised learning, the hyperplanes are determined so that data pairs
 with a common label are separated by small Hamming distances,
 and data pairs that do not have any common label are separated by large Hamming distances.


A single hyperplane can be specified by specifying its normal vector. 
Since the length of the normal vector specifying a hyperplane is immaterial,
 these lengths are chosen so that the configuration space of normal vectors
 corresponds to an $N-1$-dimensional hypersphere $S^{N-1}$. 
When distinguishing between $B$ hyperplanes, the configuration space of the hyperplanes is $(S^{N-1})^B$.


In one hashing method, the $B$ hyperplanes are set randomly~\cite{LSH_RandomProjection}.
In the following, this is referred to as the LSH method.


Other references such as~\cite{MLH,SIMBA_KN,LSH_FS1,LSH_FS2,PCAH} describe hashing methods
 that use hyperplanes. 
In particular, MLH~\cite{MLH} and S-LSH~\cite{SIMBA_KN} are described
 in subsections~\ref{subsec_MLH} and~\ref{subsec_slsh}.

\subsection{Minimal loss hashing}
\label{subsec_MLH}


One existing learning method is Minimal Loss Hashing (MLH)~\cite{MLH}.
In MLH, the aim is to minimize an empirical loss function on $(S^{N-1} )^B$. 
However, the empirical loss function is discontinuous, so it is not possible to use learning methods
 based on gradients.
Therefore, the empirical loss is replaced by a differentiable upper bound function $g$, 
 and gradients are used to minimize $g$ instead. 
The point that gives the minimum value determines the coordinates of the $B$ learned hyperplanes. 
Function $g$ has several parameters that need to be adjusted. 
Some of these parameters are dependent on the data pairs used for training. 
All the data pairs for learning are chosen at random.

\subsection{Locality-sensitive hashing with margin based feature selection}
\label{subsec_slsh}


In this subsection, we describe the concept of an existing learning method called
 locality-sensitive hashing with margin based feature selection (S-LSH)~\cite{SIMBA_KN}. 
In S-LSH, the normal vectors of hyperplanes are not used directly for learning. 
$\tilde{B}$ hyperplanes ($\tilde{B} > B$) are randomly provided. 
A degree of importance is allocated to each hyperplane,
 and these degrees of importance are calculated by learning. 
The degrees of importance are arranged in descending order,
 and the topmost $B$ normal vectors are selected. 
The distance calculations during learning are performed using weighted Hamming distances. 
Two types of data pairs are used during learning. 
The learning data pairs are selected as follows. 
A feature vector $a$ is randomly selected from the learning data. 
The first type of data pair consists of the pair $(a,b)$, where $b$ is the feature vector
 with the smallest weighted Hamming distance in the data set that has a common label as $a$. 
The second type of data pair consists of the pair $(a,c)$,
 where $c$ is the feature vector with the smallest weighted Hamming distance in the data set
 that does not have any common label as $a$.


S-LSH has been shown to have good learning performance in many data sets~\cite{SIMBA_KN}. 
It is particularly effective for learning in cases where there are many labels,
 and data with the same label has little cardinality.

\section{The proposed method}
\label{sec_proposal}
\subsection{Motivation}
\label{subsec_motivation}


To perform a high-speed similarity search that accurately represents the latent similarities
 of unstructured data, we consider performing learning with a greater number of bits $B$. 
In learning, the normal vectors of the hyperplanes are determined by making
 the Hamming distances smaller between data pairs with a common label and larger between data pairs
 that do not have any common label. 
In the following, we will refer to a data pair with a common label as a ``positive pair'',
 and to a data pair that do not have any common label as a ``negative pair''.


The configuration space of a set of $B$ hyperplanes is $(S^{N-1})^B$. 
When there is an evaluation function $U'$ on $(S^{N-1})^B$ that has the following performance,
 learning a set of hyperplanes can be regarded as an optimization problem that globally maximizes $U'$.
The argument of $U'$ is the arrangement of multiple hyperplanes.
Each hyperplane divides the feature space $V$ into two regions. 
The value of $U'$ increases
 as the number of positive pairs whose feature vectors are in a same region 
 and negative pairs whose feature vectors are in different regions increase.
In most cases, a function $U'$ having this property is thought to have multiple local maxima. 
When $B$ is large, the dimension of $(S^{N-1})^B$ increases and it becomes harder to solve
 the optimization problem. 
Since we are concerned here with hashing using hyperplanes, the values of function $U'$ can also be discrete. 
Therefore, it is unnatural to require continuity of the $U'$ configuration space $(S^{N-1})^B$. 
Since $U'$ is not necessarily differentiable, it cannot be
 globally maximized by methods that use the gradient of $U'$.


Instead of solving an optimization problem in $(S^{N-1})^B$,
 we can consider a method where optimization problems in $S^{N-1}$ are solved $B$ times,
 and these solutions are bundled together. 
That is, instead of learning a set of $B$ hyperplanes, the individual hyperplanes are
 separately learned and the results are bundled together.
However, if we obtain $B$ solutions to the optimization problem in $S^{N-1}$,
 then the performance is severely impaired for the following reason. 
Consider an evaluation function $U$ on $S^{N-1}$. 
Assume that the points $S^{N-1}$ where the value of $U$ is larger correspond to a good hyperplane. 
That is, we assume the following property.
When the feature space $V$ is partitioned into
 two regions by a single hyperplane,
 the value of $U$ increases
 as the number of positive pairs whose feature vectors are in a same region 
 and negative pairs whose feature vectors are in the different regions increase.
A specific example of an evaluation function is shown in subsection~\ref{subsec_EvaluationFunction}. 
We will assume that the evaluation function $U$ has a global maximum value on $p_* \in S^{N-1}$. 
If all the hyperplanes exist in $p_*$, then they are all degenerate. 
In this case, the feature space $V$ is only divided into two regions,
 and there are only two types of representative bit string. 
Clearly it would not be possible to capture the features of unstructured data with these bit strings. 
For this reason, when we consider bundling together the learning results of individual hyperplanes,
 it can be said that individual hyperplanes are not necessarily learned by finding the point
 where the evaluation function $U$ on $S^{N-1}$ is maximized. 
Therefore, in the following we consider finding $B$ points where the evaluation function $U$ has
 a local maximum value in $S^{N-1}$ when performing learning with individual hyperplanes. 
Multiple hyperplanes are learned by bundling these together. 
Here, we must ensure that the multiple hyperplanes are not oriented in the same direction.


In the remainder of this section, we describe the proposed method,
 which is a hyperplane normal vector learning method (referred to below as M-LSH) based on
 the Markov Chain Monte Carlo method. 
At the same time, we consider and discuss a number of evaluation functions
 (which have a strong influence on the performance of M-LSH learning),
 and data pair sub-sampling methods.

\subsection{Learning hyperplanes with the Markov chain Monte Carlo method}


In this section, we describe the proposed method. 
Our proposed method is a supervised learning method for hyperplanes using MCMC. 
The aim of this learning method is to probabilistically determine the point
 where the evaluation function $U$ reaches its local maximum value. 
The advantage of this method is that it does not require a differentiable evaluation function. 
Its disadvantage is that because it uses a Monte Carlo method,
 the point where the evaluation function is locally maximized cannot be determined with perfect accuracy. 
However, from the properties of MCMC, the learned results are highly likely to be close
 to the point where the evaluation function is locally maximized.
The probability that a particle is in such a place
 is high so that the local maximum value is high, and the peak is sharp.


In the following, we will assume that the evaluation function $U$ is a function
 whose values are positive and are bounded on $S^{N-1}$. 
Also, the details of the evaluation function are assumed to depend on the training data pairs. 
Specific examples of $U$ are given in subsection~\ref{subsec_EvaluationFunction}.


Consider a particle on $S^{N-1}$ whose position equates to the normal vector of a hyperplane. 
If $\tilde{U}:=-U$ is regarded as the potential energy, then to obtain the local minimum value of $\tilde{U}$,
 we need to consider the motion of dissipative particles. 
However, since $U$ is generally not differentiable, we cannot use optimization methods based on gradients,
 that is, continuous particle motions. 
Therefore we can consider obtaining a minimum solution by a random walk method.


We regard $U$ as the probability density function of $S^{N-1}$ (except for a normalization constant),
 and use MCMC to evaluate the temporal evolution of particles.
This method is our proposed M-LSH method.


We use the Metropolis-Hastings algorithm for MCMC~\cite{Hastings_1970}. 
For the proposed density function, we use the normal distribution. 
In M-LSH, particles perform random walks a fixed number of times. 
This is the temporal evolution of the particles. 
We refer to this temporal evolution as a single batch process. 
Since the details of the evaluation function are determined by deciding on the training data pairs,
 the handling of the training data pairs may lead to incidental local maximum values of the evaluation function. 
To prevent the particles from becoming trapped at this sort of point,
 batch processing is performed a number of times,
 and the learning data pairs are replaced for each batch process.


By performing learning with multiple hyperplanes,
 we obtain multiple points where the evaluation function is locally maximized. 
As described in subsection~\ref{subsec_motivation}, it is necessary to prevent the learning
 of points where multiple hyperplanes produce the same local maximum value. 
In M-LSH, this issue is resolved by using the following method. 
MCMC exhibits a property whereby particles tend to accumulate at places where the probability
 density function is locally maximized. 
The sharper the peak in the evaluation function close to the local maximum value,
 the more intense this trend becomes. 
In most cases, since $U$ is multimodal,
 making its peaks sharper and randomly setting the initial positions of the particles
 will cause the particles to collect at peaks close to their initial positions. 
Therefore, we can prevent the particles from all moving towards the same point.

\begin{figure}[htb]
	\begin{center}
	\includegraphics[scale=0.4]{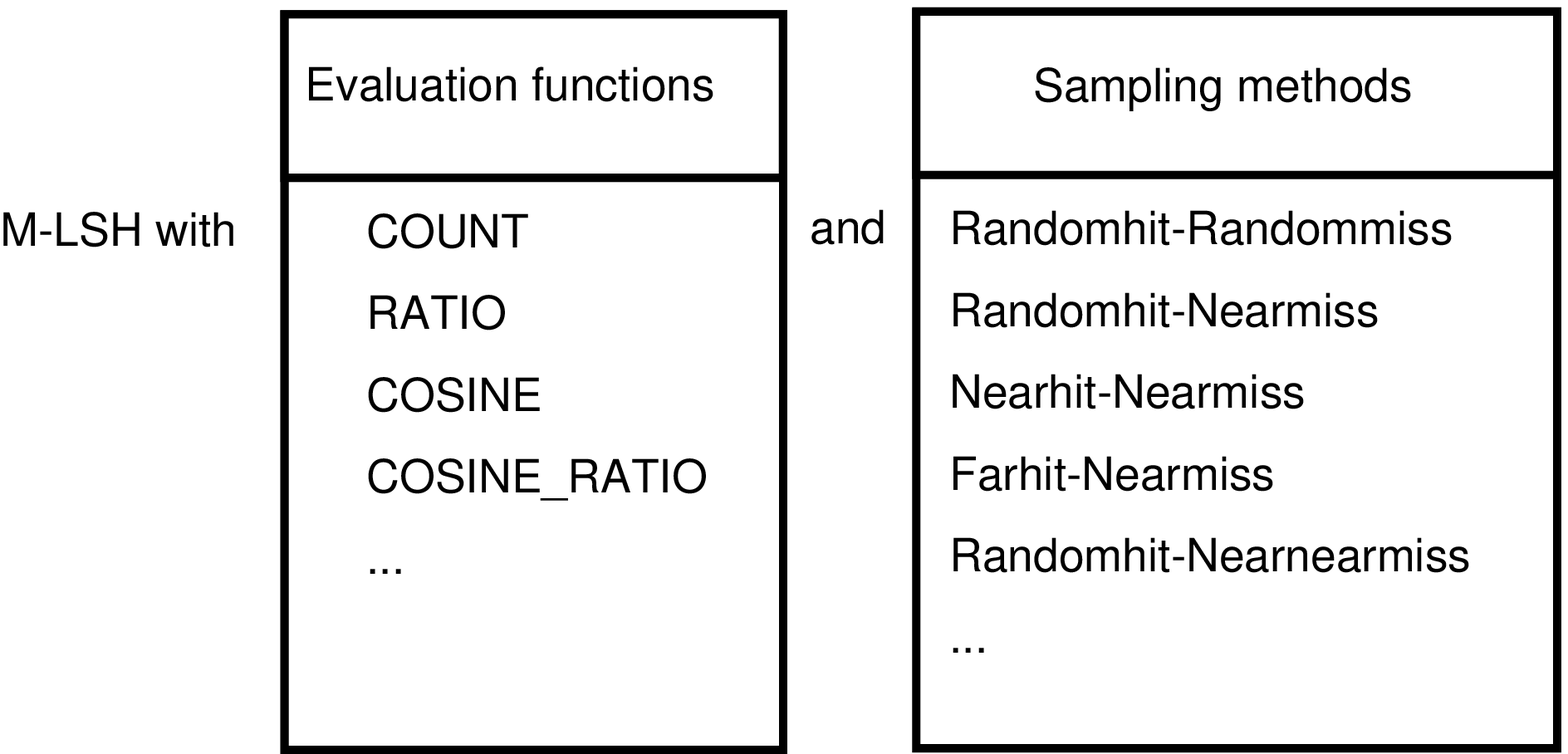}
	\end{center}
 	\caption{M-LSH variants. The number of variant is the number of combinations of
		evaluation functions and sampling methods }
	\label{fig_M-LSHCombination}
\end{figure}

Many variants of M-LSH can be considered. 
These variants can be obtained by combining the evaluation function $U$ with sampling methods
 for training data pairs that determine the details of the evaluation function. 
Table~\ref{fig_M-LSHCombination} lists these combinations. 
Each of these items is described below in subsections~\ref{subsec_EvaluationFunction} and~\ref{subsec_Sampling}.

\subsection{Evaluation function}
\label{subsec_EvaluationFunction}


When learning is performed by M-LSH, the type of evaluation function must be determined. 
A number of possible evaluation function types are considered below.
First we will introduce some nomenclature. 
$PP$ denotes the set of all given positive pairs, and $NP$ denotes the set of all given negative pairs. 
The angles subtended by the two feature vectors of a pair $p$ relative to the normal vector
 of a hyperplane are $\theta_1(p)$ and $\theta_2(p)$, respectively. 
The following subsets are defined.
\begin{eqnarray}
	PP_+ &:=& \left\{
				p\in PP |
				\cos(\theta_1(p))*\cos(\theta_2(p)) > 0
			\right\} ,\\
	NP_{-} &:=& \left\{
				p\in NP |
				\cos(\theta_1(p))*\cos(\theta_2(p)) < 0
			\right\} .
\end{eqnarray}
In the following, the cardinality of a set $A$ is denoted by $\#A$.


In the following formulas, we assume an evaluation function $U = \exp(x /T)$
 that uses an enumerated value $x$. Here, $T=1$.
\begin{description}
	\item[COUNT]
		\begin{eqnarray}
			x = \#PP_+ + \#NP_- .
		\end{eqnarray}
	\item[RATIO]
		\begin{eqnarray}
			x = \frac{\#PP_+}{\#PP} + \frac{\#NP_-}{\#NP} .
		\end{eqnarray}
	\item[COSINE]
		\begin{eqnarray}
			x &=& 
			\sum_{p\in PP} |\cos(\theta_1(p))+\cos(\theta_2(p))| \nonumber\\
			&&+\sum_{p\in NP} |\cos(\theta_1(p))-\cos(\theta_2(p))| .
		\end{eqnarray}
 	\item[COSINE\_RATIO]
 		\begin{eqnarray}
			x &=& 
			\frac{1}{\#PP}\sum_{p\in PP} |\cos(\theta_1(p))+\cos(\theta_2(p))| \nonumber\\
			&&+\frac{1}{\#NP}\sum_{p\in NP} |\cos(\theta_1(p))-\cos(\theta_2(p))| .
		\end{eqnarray}
\end{description}


If $-x$ and $T$ are regarded as the particle's energy and temperature respectively,
 then $U$ can be regarded as a Boltzmann weight. 
From this perspective, the low-temperature limit is where $T$ is zero,
 and the high-temperature limit is where $T$ is $\infty$.

\subsection{Sampling method for training data}
\label{subsec_Sampling}


The evaluation function defined in subsection~\ref{subsec_EvaluationFunction} must include both $PP$ and $NP$. 
$PP$ and $NP$ can be determined by considering all combinations of the training data. 
However, the way in which the data pairs are obtained means that the number of pairs is only half
 the square of the number of training data items. 
When the cardinality of $PP$ and $NP$ is large, it can take a long time to calculate the evaluation function.


Therefore in this subsection we consider a number of different selection methods for $PP$ and $NP$,
 and we discuss their advantages and disadvantages. 
Here, we will use the term ``distance'' to refer to L2 distance unless otherwise noted.


We will also use the following nomenclature. 
$L$ is the set of all the training data. 
$L_a$ represents a data set having a common label as an element $a\in L$,
 and $L_a^c$ represents $L \setminus L_a$.
The distance between two elements $a,b\in L$ is denoted by $dist(a,b)$.

%
%

We will start by discussing the selection method for $NP$. We will consider the following sampling method.
\begin{description}
	\item[Randommiss]\mbox{}\\
	After $a\in L$ has been randomly selected, $b\in L_a^c$ is randomly selected to form a negative pair $(a,b)$.
	
	\item[Nearmiss]\mbox{}\\
	After $a\in L$ has been randomly selected, a negative pair $(a,b)$ is
		formed such that $b := \arg\min_{c\in L_a^c}(dist(a,c))$.
	
	\item[Boundarymiss]\mbox{}\\
	After $a\in L$ has been randomly selected, a negative pair $(a',b)$ is
		formed such that $b := \arg\min_{c\in L_a^c}(dist(a,c)) $ and
		$a' := \arg\min_{c\in L_a\cap L_b^c} (dist(b, c))$.
\end{description}

\begin{figure*}[htb]
	\begin{center}
 	\includegraphics[scale=0.10]{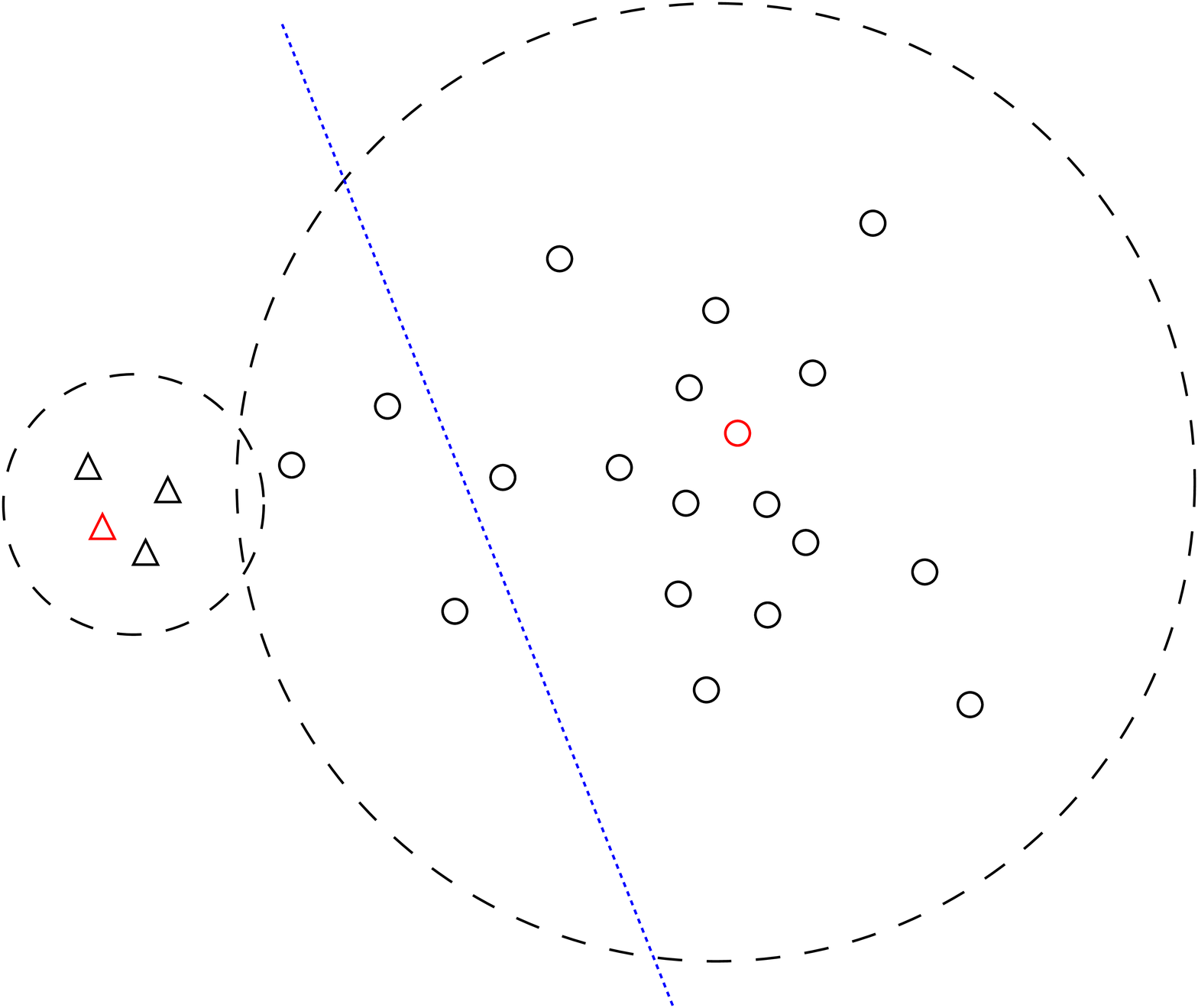}
 	\hspace{5mm}
 	\includegraphics[scale=0.10]{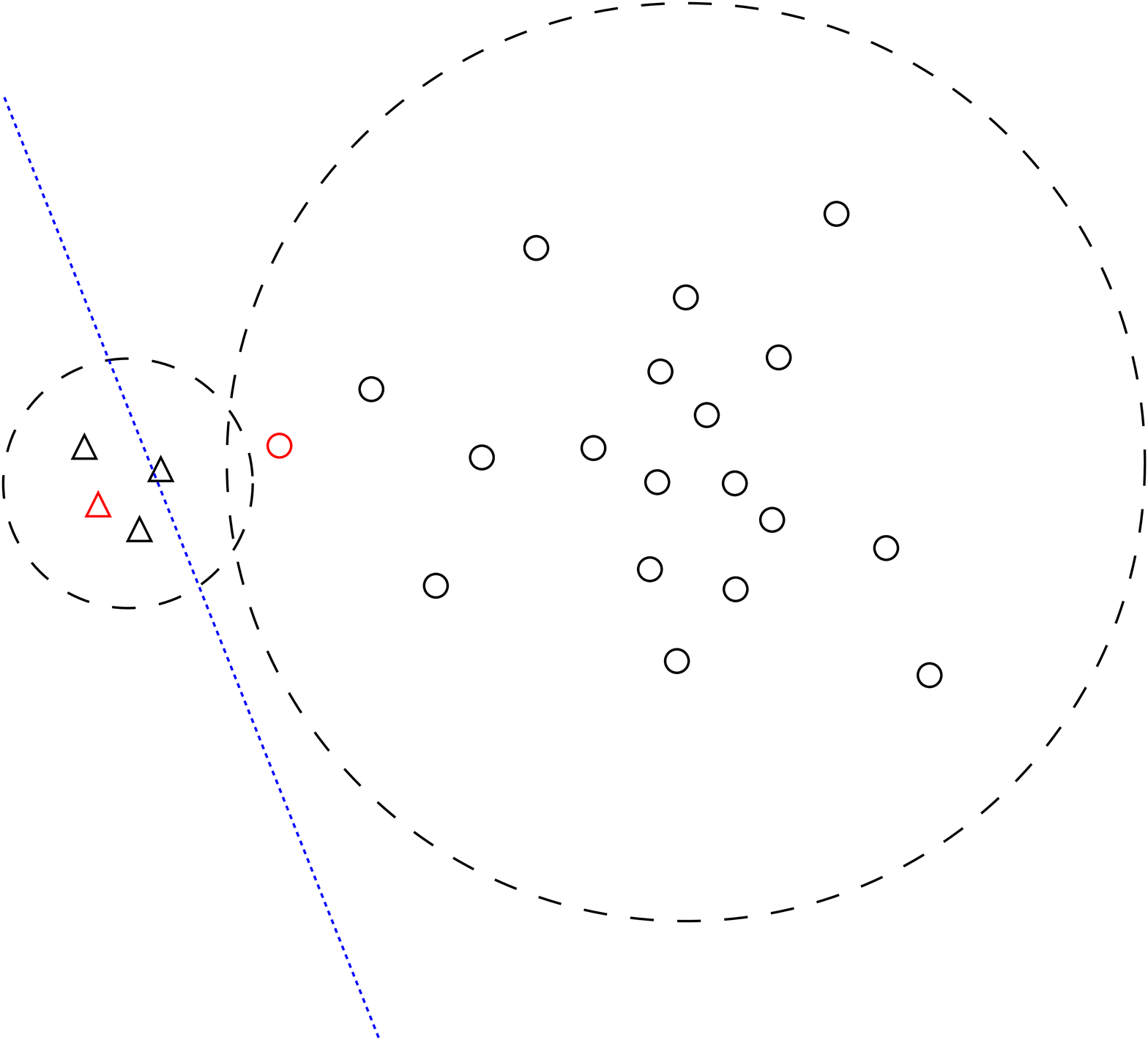}
 	\hspace{5mm}
 	\includegraphics[scale=0.10]{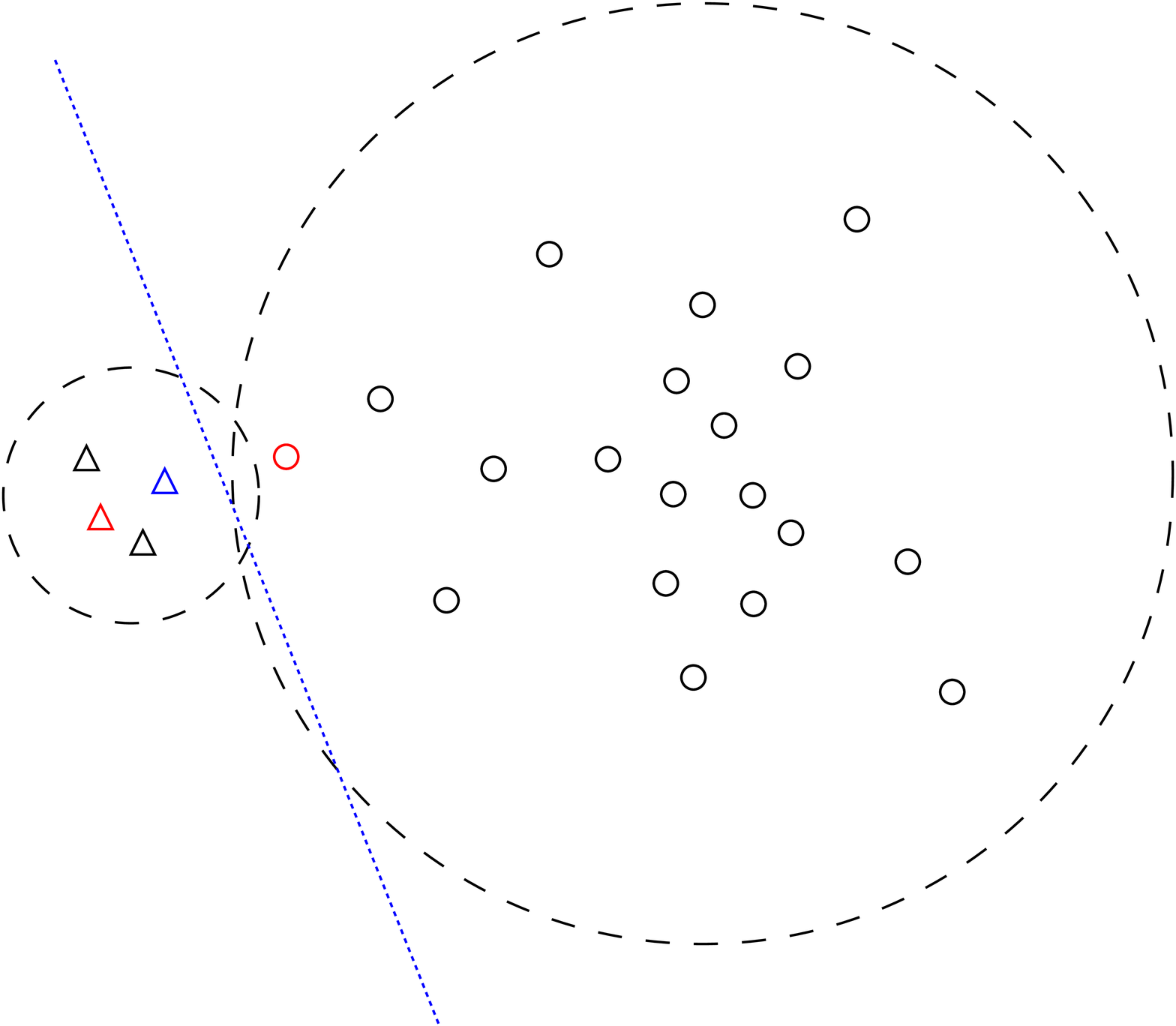}
 	\end{center}
 	\caption{Selection methods for negative pairs.
	(From left to right) Randommiss, Nearmiss, Boundarymiss.
	The red triangles, red circles and blue triangles correspond to element $a$,
	 element $b$ and element $a'$ respectively. 
	The blue dotted line represents the hyperplane expected to be obtained by learning.
		}
 	\label{fig_NPsampling}
 \end{figure*}

Since Boundarymiss as defined above is a new sampling method,
 we will describe it in more detail here. 
Consider two elements $a, b\in L$ that do not have any common label and $L_a \cap L_b \not= \emptyset$. 
Since the labels applied to unstructured data can be of more than one type,
 this sort of situation can occur frequently. 
Since the distributions of $L_a$ and $L_b$ are overlapping,
 it is not possible to obtain a hyperplane that separates them completely. 
If we are allowed to bisect $L_a \cap L_b$ with a hyperplane,
 then it may also be possible to separate the difference sets $L_a \setminus L_b$ and $L_b \setminus L_a$. 
To learn a hyperplane that bisects $L_a \cap L_b$, we can form a negative pair by selecting
 one data item from each of $L_a \setminus L_b$ and $L_b \setminus L_a$. 
Boundarymiss is one of the ways in which negative pairs of this sort can be made. 
Furthermore, Boundarymiss is expected to lie close to the boundary
 between $L_a \setminus L_b$ and $L_b \setminus L_a$. 
Please refer to Fig.~\ref{fig_NPsampling}.


In Randommiss sampling, there is a high possibility of selecting a pair comprising an element close
 to the center of gravity of $L_a$ and an element close to the center of gravity of $L_a^c$. 
This makes it easier to learn a hyperplane that separates the center of gravity of $L_a$ from the center
 of gravity of $L_a^c$. 
When $L_a^c$ is distributed over a broader region than $L_a$,
 it is expected that the resulting hyperplane will be deviated from the boundary of $L_a$ and $L_a^c$. 
In particular, when the number of labels applied to the training data is large and the sets of each label
 have similar cardinality, the distribution of $L_a^c$ tends to become broader than that of $L_a$,
 so the tendency for the learned hyperplanes to be separated from the boundary is thought
 to become more pronounced as the number of labels increases. 
Figure~\ref{fig_NPsampling} shows some typical data pairs obtained by Randommiss sampling,
 and the hyperplanes learned from these pairs. 
The dotted circles in these figures show the approximate regions over which these sets are distributed.


In Nearmiss sampling, an element selected from $L_a^c$ lies close to the boundary of $L_a$ and $L_a^c$,
 so it is possible to avoid the above drawback of the Randommiss sampling method.
However, when $L_a$ is distributed over a wide region,
 there is a greater likelihood of $a\in L$ being deviated from the boundary between $L_a$ and $L_a^c$. 
Figure~\ref{fig_NPsampling} shows some typical data pairs obtained by Nearmiss sampling,
 and the hyperplanes learned from these pairs.


When Boundarymiss sampling is performed, it can compensate for the abovementioned drawbacks
 of the Nearmiss method, but is liable to choose data pairs that are separated by smaller
 distances and is therefore more susceptible to noise in the data.


%
%

We will now describe the selection method for $PP$. 
For the reasons discussed below, it is better to consider the handling of positive pairs
 in terms of applying corrections to the discriminant planes used for the discrimination of negative pairs. 
For example, if we learn with only positive pairs,
 because the hyperplanes should not separate feature vectors in each positive pair,
 the hyperplanes may stay away from all the feature vectors. 
In this case, all the bit strings of the training data will be identical,
 making it impossible to separate the feature vectors. 
We therefore consider positive pairs to have the role of preventing $L_a$ from becoming separated
 by the hyperplane. 
We will consider the following sampling methods for positive pairs.
\begin{description}
	\item[Randomhit]\mbox{}\\
After $a\in L$ has been randomly selected, $b\in L_a$ is randomly selected to form a positive pair $(a,b)$.
	
	\item[Nearhit]\mbox{}\\
After $a\in L$ has been randomly selected, a positive pair $(a,b)$ is
 formed such that $b := \arg\min_{c\in L_a}(dist(a,c))$.
	
	\item[Farhit]\mbox{}\\
After $a\in L$ has been randomly selected, a positive pair $(a,b)$ is
 formed such that $b := \arg\max_{c\in L_a}(dist(a,c))$.
\end{description}


We consider a data set whose element have a single label.
In this case, it is considered that Farhit sampling frail against outliers. 
It is thought that Randomhit sampling is robust against outliers. 
Nearhit sampling is expected to have poor performance because it is not possible to prevent data other
 than the selected data pair in $L_a$ from being arranged in different directions of the hyperplane. 
It is thought that this performance degradation is particularly severe when there are many data items
 with the same label.

\begin{figure}[htb]
	\begin{center}
 	\includegraphics[scale=0.20]{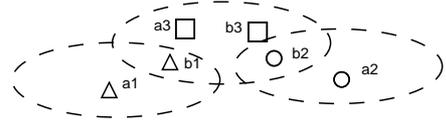}
 	\end{center}
\caption{Schematic illustration of overlapping data pairs
}
\label{fig_overlappingDataPair}
\end{figure}

We consider a data set whose element have an arbitrary number of labels.
Here, we consider the case where there are three positive pairs $(a_1,b_1),(a_2,b_2),(a_3, b_3) \in PP$,
 such that $a_1 \notin L_{a_2} \wedge b_1 \in L_{b_2} \wedge a_3, b_3 \in L_{b_2}$. 
In particular, when $a_3$ and $b_3$ are close to $b_1$ and $b_2$ respectively,
 we shall refer to these data pairs as overlapping data pairs. 
This is summarized in Fig.~\ref{fig_overlappingDataPair}.
When learning is performed in this case, it becomes difficult to separate $L_{a_1}$ and
 $L_{a_2}$ from $L_{b_2}$.
As the number of sampling pairs increases, it is thought that overlapping data pairs will become more common. 
It is therefore expected that Farhit sampling and Randomhit sampling will cause the performance to become worse. 
In the case of Nearhit sampling, since $a_3$ and $b_3$ are less likely to be close to $b_1$ and $b_2$,
 it is thought that performance degradation will be less likely to occur.


Based on this reasoning, there are as many possible sampling methods as
 there are combinations of positive pair and negative pair sampling methods. 
The sampling methods that we actually evaluated and compared are as follows\footnote{
We believe that this choice is natural.}
.
\begin{itemize}
	\item Randomhit-Randommiss
	\item Randomhit-Nearmiss
	\item Nearhit-Nearmiss
	\item Farhit-Nearmiss
	\item Randomhit-Boundarymiss
\end{itemize}

\section{Experiments and evaluation}
\label{sec_experimentsAndResult}


We performed experiments to measure the effects of the proposed method on a number of different data sets. 
In these experiments, supervised learning was performed on data that had already been labeled. 
The data labels used in these experiments are all known. 
When the search results are obtained, the Hamming distance between the query and data
 in the database is calculated, and the top search results are ordered in ascending order of distance. 
The acquisition rate is defined for this purpose as follows.
\begin{eqnarray}
	\mbox{Acquisition} := \frac{\mbox{\small Number of data acquired by search}}
					{\mbox{\small Total number of data searched}}.
\end{eqnarray}
To evaluate the performance, we used the precision rate and recall rate as defined below.
\begin{eqnarray}
	\mbox{Precision}
		&:=& \frac{\begin{minipage}[htb]{50mm}{\small Number of data items with a common label
	as the query for which search results were obtained} \end{minipage}}
			{\begin{minipage}[htb]{50mm}{\small Number of data items for which search results were obtained}\end{minipage}} ,\\
			&&\nonumber \\
	\mbox{Recall}
		&:=& \frac{\begin{minipage}[htb]{50mm}{\small Number of data items with a common label
	as the query for which search results were obtained}\end{minipage}}
			{\begin{minipage}[htb]{50mm}{\small Number of data items with a common as the query among all relevant items}\end{minipage}} .
\end{eqnarray}
A recall-precision curve shows the variation of recall rate and precision rate
 with changes in the acquisition rate. 
Better search performance is indicated by a recall rate and precision rate with values closer to 1.


In the experiments, by way of reference,
 we also calculated the precision rate and recall rate in similarity searches based on L2 distances
 using the original feature vectors.


The remainder of this section is structured as follows. 
First, we confirm the benefits of M-LSH on learning with an artificial data set that we prepared. 
We then evaluate the performance of the evaluation functions and sampling methods considered
 in sections~\ref{subsec_EvaluationFunction} and~\ref{subsec_Sampling}. 
For this performance evaluation, we used actual data sets instead of our prepared data set. 
Finally we show how our proposed method differs from existing learning methods.

\subsection{Experiments with an artificial data set}


Using an artificial data set, we confirmed the effects of M-LSH on learning. 
This artificial data set consists of 300 data items sampled from
 a three-dimensional standard normal distribution. 
With the axes labeled as x, y and z, we classified the data items into two classes according
 to whether the x component was positive or nonpositive. 
As can be seen from the way in which the data is labeled,
 we desire a hyperplane whose normal vector $\vec{n}$ is $\vec{n} = (\pm 1, 0,0)$.

\begin{figure*}[htb]
	\begin{center}
 	\includegraphics[scale=0.5]{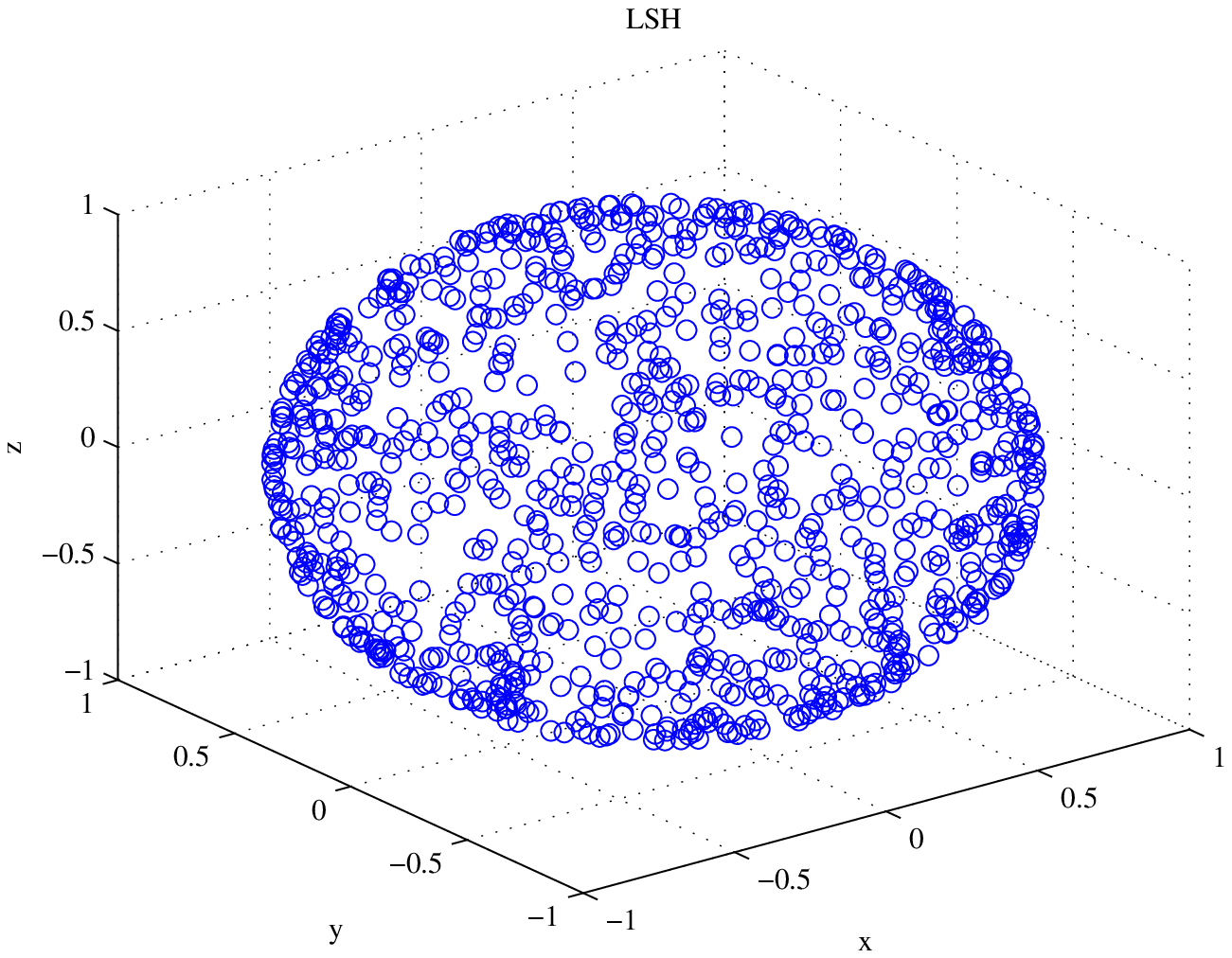}
 	\includegraphics[scale=0.5]{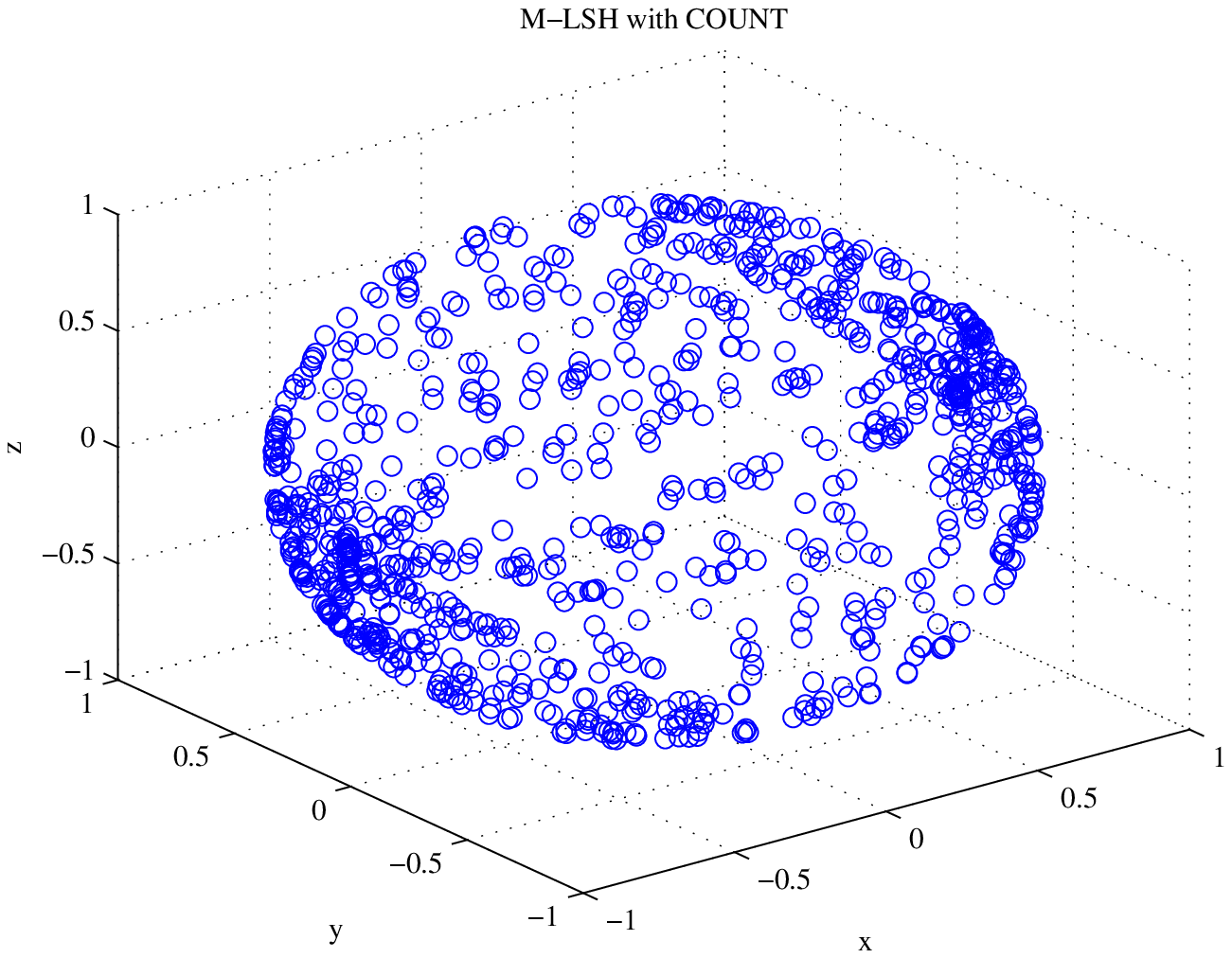}
 	\\
 	\includegraphics[scale=0.50]{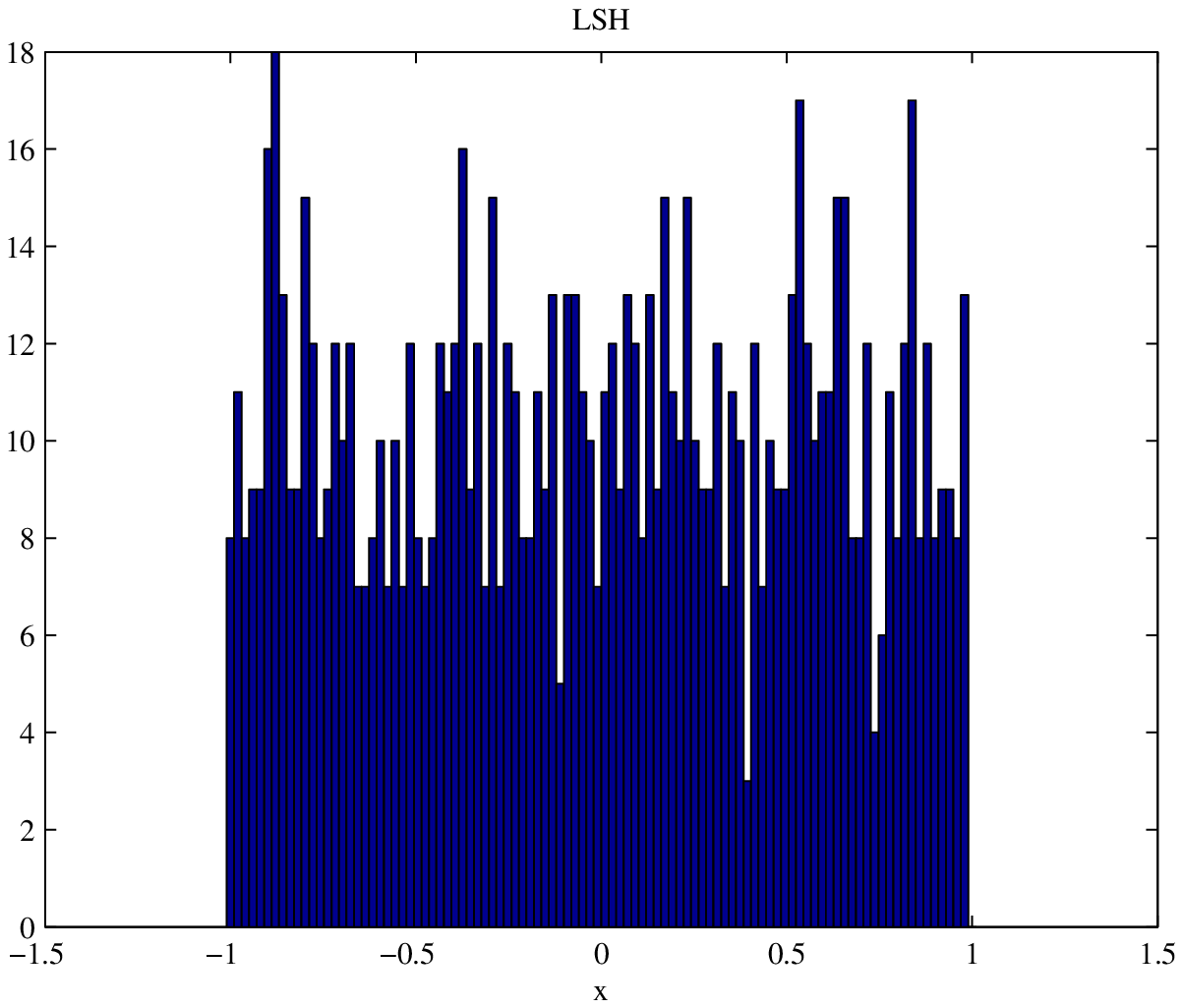}
	\includegraphics[scale=0.50]{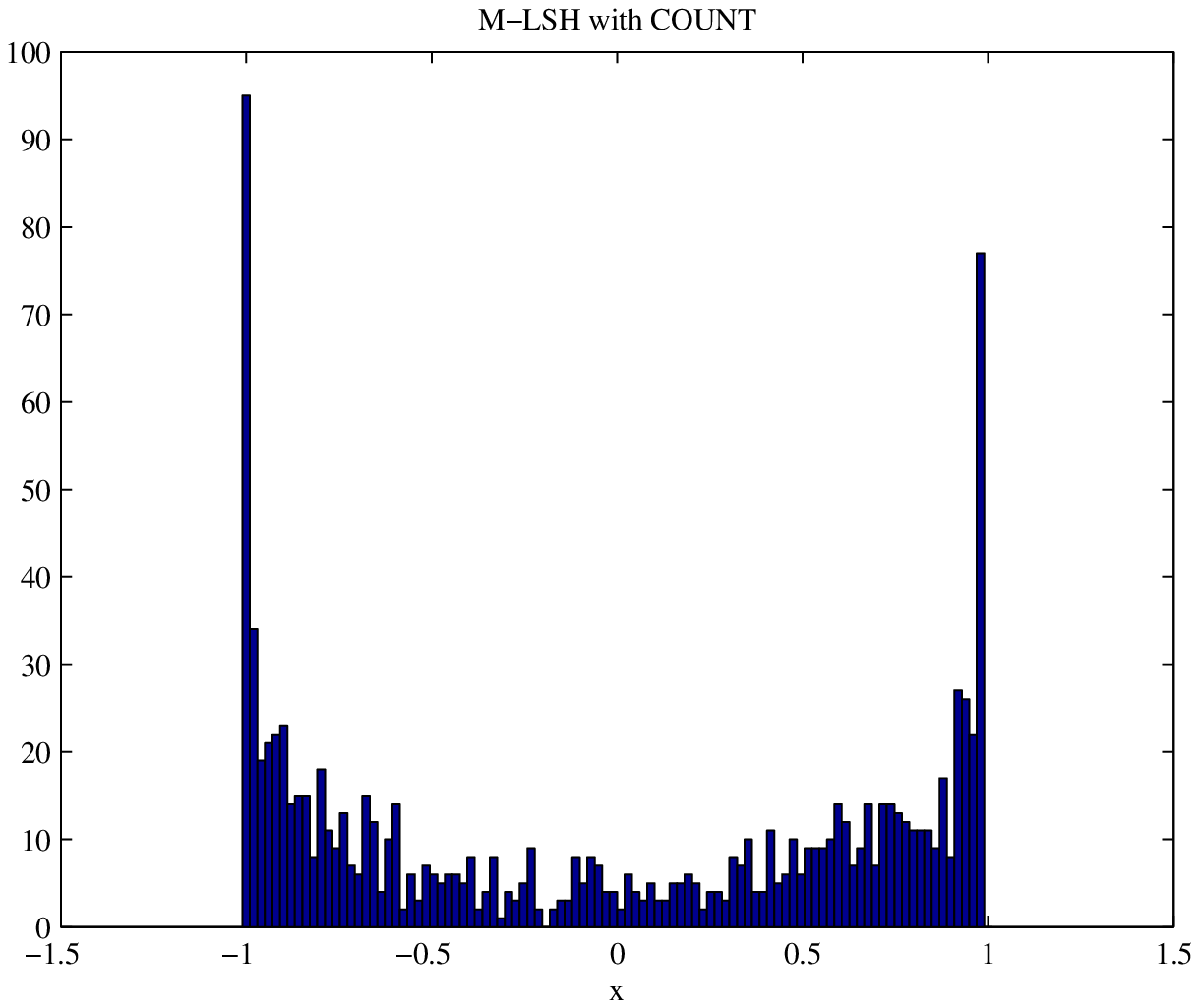}
 	\end{center}
\caption{LSH and M-LSH learning results with artificial data set.
 Top left: Scatter diagram of normal vectors learned by LSH; Bottom left: Histogram of $x$ components. 
Top right: Scatter diagram of normal vectors learned by M-LSH; Bottom right: Histogram of $x$ components.
}
\label{fig_M-LSHEffect}
\end{figure*}

Figure~\ref{fig_M-LSHEffect} shows the effects of LSH and M-LSH on learning with a bit string
 length of 1,024. 
The parameters of learning with M-LSH were as follows: number of processing batches: 5,
 number of temporal evolution steps in batch processing: 100,
 number of data pairs used for learning in each batch process: 2,000,
 number of evaluation functions used during learning: COUNT,
 sampling method: Randomhit-Randommiss, with equal numbers of positive and negative pairs.


\begin{figure}[tb]
	\begin{center}
	\includegraphics[scale=0.35]{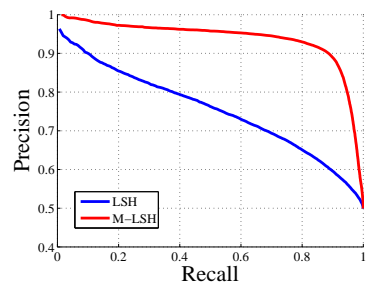}
 	\end{center}
\caption{Precision rate and recall rate curves of LSH and M-LSH learning with artificial data set.
}
\label{fig_TrivialXPR}
\end{figure}

From Fig.~\ref{fig_M-LSHEffect}, we can see the following.
From the scatter diagram and $x$ component histogram of the normal vectors obtained by LSH,
 we can see that the normal vectors are uniformly distributed on a two-dimensional sphere. 
From the scatter diagram and $x$ component histogram of the normal vectors obtained by M-LSH,
 we can see that most of the normal vectors are distributed in the vicinity of $\vec{n}=(\pm 1, 0,0)$. 
Figure~\ref{fig_TrivialXPR} shows the precision rates and recall rates of LSH and M-LSH.
As expected from the distribution of normal vectors, Fig.~\ref{fig_TrivialXPR} shows that M-LSH has
 a positive effect on learning.


\subsection{Experimental data}


In this subsection, we describe the experimental data used in the performance evaluations
 performed in subsections~\ref{subsec_resultEvalFunc} and~\ref{subsec_resultSampling}.






The experimental data was obtained from the following sources.
\begin{itemize}
\item MNIST\\
Scanned images of handwritten numerals 0--9~\cite{MNIST}. 
Each digit is stored as a $28\times28$-pixel 8-bit grayscale image,
 and is labeled with the corresponding digit 0--9. 
We used the images themselves as feature quantities. 
Therefore, the feature quantities had 784 dimensions.

\item 
Fingerprint images
\\
Fingerprint image data acquired using a fingerprint image scanner. 
The feature quantities consisted of the 4,096-dimensional Fourier spectra of these fingerprint images. 
Since fingerprints are unique to each human, the data was labeled with the names of the corresponding individuals. 
In other words, each feature vector was given just one label.
For details, see Ref.~\cite{SIMBA_KN}.

\item 
Speech features
\\
A set of 200-dimensional mel frequency cepstral coefficient (MFCC) feature quantities extracted
 from a three-hour recording of a local government assembly published on the Internet~\cite{KawasakiCouncil}. 
The query was a spoken sound acquired separately. 
In supervised learning of speech, the contents of the speech are normally labeled with a text transcription. 
But instead,
 we treated
 the features with the top 0.1\% shortest Euclidean distances from the queries were regarded to be in the
 same class.
Each feature vector could have multiple labels.

\item LabelMe\\
LabelMe data using 512-dimensional Gist feature quantities~\cite{GIST} extracted from image data
 published in Ref.~\cite{512dimGISTdata}. 
The labeling was applied to the dissimilarity matrices of distributed data,
 and the same labels were applied to data corresponding to the topmost 50 rows of data in each row. 
Each feature vector can have multiple labels.
\end{itemize}
These data sets were selected with the following applications
 in mind --- MNIST: handwritten number recognition, Fingerprint images: biometric identification,
 Speech features: speech recognition, LabelMe: automatic image classification.


The quality of data used in the experiments is summarized in Table~\ref{table_NumbersForExperiments}. 
Here, we envisaged performing searches on data recorded in a database,
 with the data divided into three pairwise disjoint sets: a data set used for learning,
 the searched data set, and a data set for queries. 
The learning performance varied widely depending on the number of labels in the data set
 and on the cardinality of data sets having a common label. 
However, since the data sets were not all given unique labels,
 it is not possible to give a na\"{i}ve definition of the label numbers. 
We therefore reasoned as follows. 
For the data actually used for learning, the average value of the number of data items having 
 a common label as one item of data is regarded as the rough cardinality of the sets for each label. 
The rough number of labels is then calculated by dividing
 the number of data items actually used for learning
 by the rough cardinality of the sets for each label. 
This information is summarized in Table~\ref{table_NumbersForExperiments}.


Since data generally contains noise, noise reduction must be performed.
Prior to the experiments, we subjected all the data to the following noise reduction processes. 
These processes are widely used as noise reduction methods. 
The feature quantities of the data are higher-dimensional data. 
Depending on the data set, each component of the data may be expressed in different units. 
Unless the feature vectors are made dimensionless, they cannot be used for the calculation
 of distances or angles. 
We therefore subjected the data to an affine transformation so that the average value of
 each component of the feature vector of the learning data became 0, and the standard deviation
 of the learning data became 1. 
We also performed a principal component analysis for the learning data. 
This was done by finding the subspace with a cumulative contribution rate of over 80\%,
 and mapping all data to this space. 
After the above noise reduction process, we performed learning and search tests. 
Since all the methods were evaluated using data that had been subjected to this noise reduction process,
 this noise reduction process had no effect on the performance of each method.


\begin{table*}[htb]
	\begin{center}
	\caption{Experimental parameters}
	\begin{tabular}{c|r|r|r|r}
		\hline
		\hline
		\backslashbox{
		Parameter
		}{Data set} & MNIST & Fingerprint & Speech & LabelMe \\
		\hline
		Number of training data items
			& 60,000 & 9,906  & 192,875 & 11,000 \\
		Number of data items for searching
		  & 5,000  & 12,138 & 192,683 & 5,500  \\
		Number of data items for queries
		  & 5,000  & 19,932 & 1,815   & 5,500  \\
		Dimension before dimensionality reduction
		  & 784    & 4096   & 200     & 512    \\
		Dimension after dimensionality reduction
		  & 149    & 276    & 30      & 20     \\
		Feature vector have unique labels
			& Yes & Yes & No& No   \\
		Approximate number of labels
		& 10 & 1300 & 2000 & 300\\
		Rough cardinality of the sets for each label
		& 6000 & 7 & 100 & 40 \\
		\hline
	\end{tabular}
	\label{table_NumbersForExperiments}
	\end{center}
\end{table*}


The parameters of the M-LSH experiments were as follows. 
Standard deviation of proposed density distribution: 0.01,
 number of processing batches: 10, number of temporal evolution steps in batch processing: 100,
 number of data pairs used for learning in each batch process: 2,000 or 20,000.

\subsection{Evaluation function performance}
\label{subsec_resultEvalFunc}


Here, we evaluate the performance of the evaluation functions cited in
 subsection~\ref{subsec_EvaluationFunction}. 
A natural choice of sampling method is the simplest Randomhit-Randommiss sampling method.
However, as was found in subsection~\ref{subsec_resultSampling},
 the Randomhit-Randommiss sampling method has poor performance.


Therefore, we instead used Randomhit-Nearmiss sampling, which is regarded as
 the next simplest sampling method after Randomhit-Randommiss sampling.

\begin{figure*}[htb]
	\begin{center}
 	\includegraphics[scale=0.35]{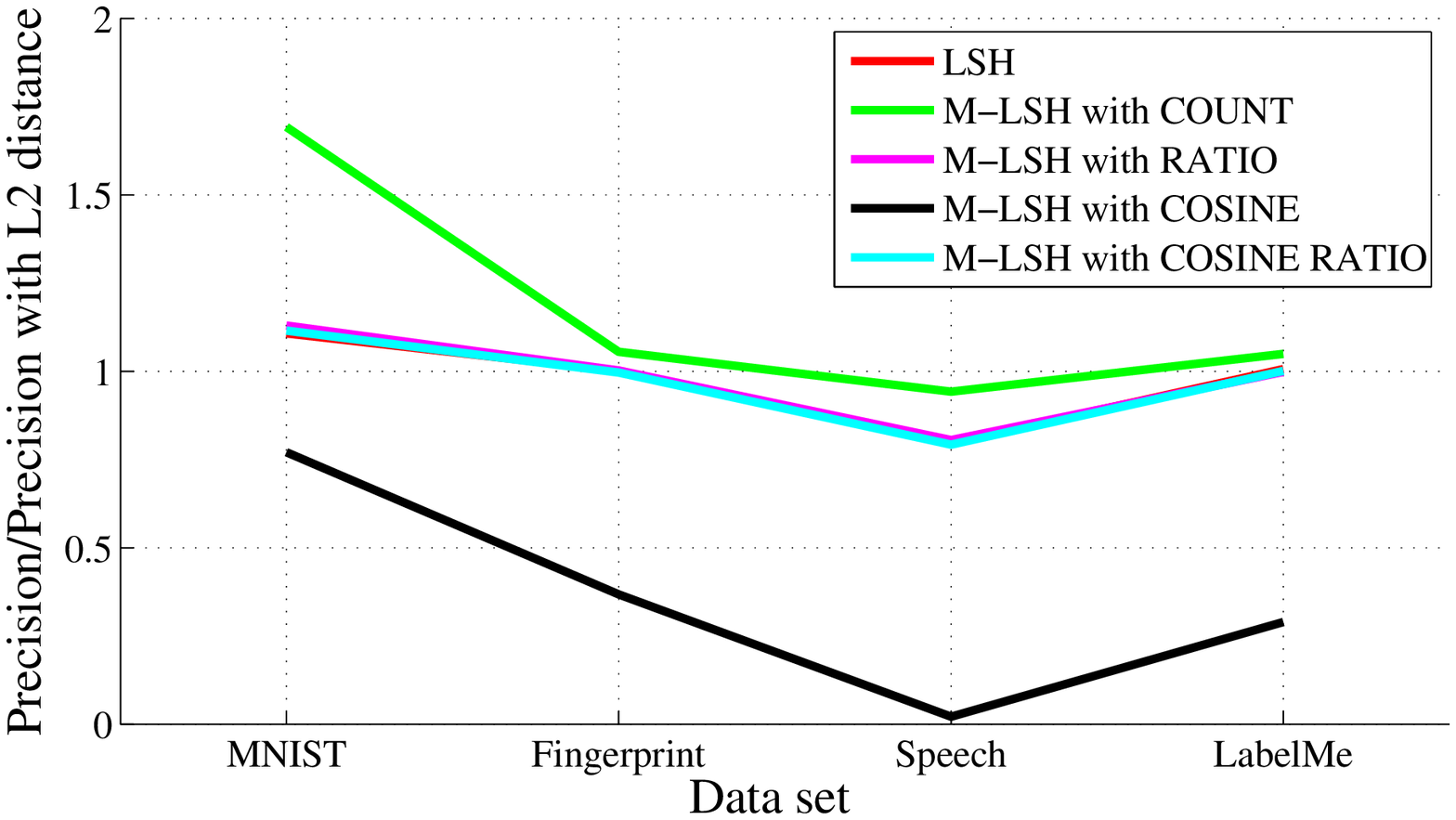}
 	\includegraphics[scale=0.35]{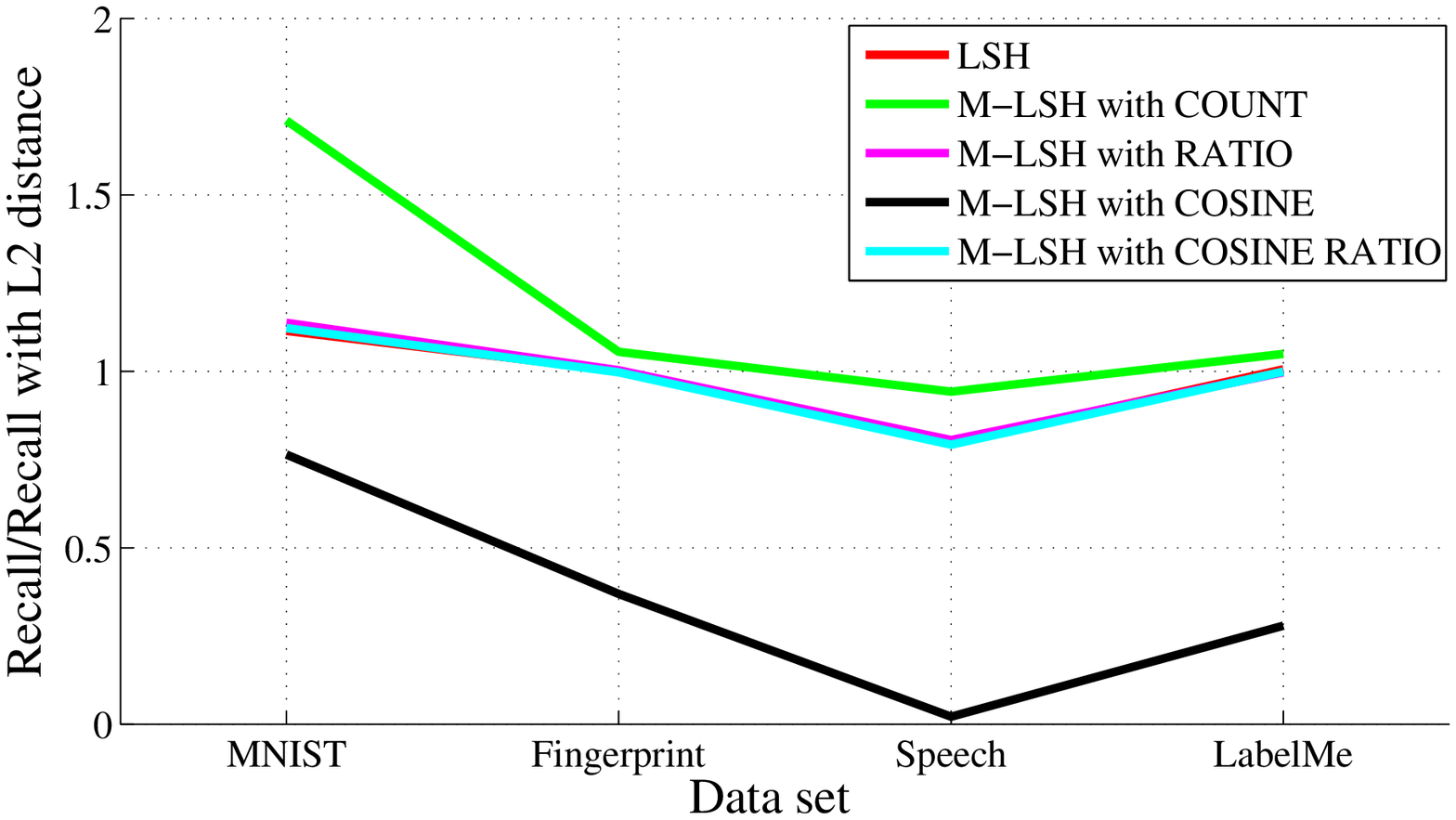}
 	\end{center}
	\caption{Performance of evaluation functions with different data types: Precision rates (left) and recall rates (right)
	}
	\label{fig_EvalFuncPR}
\end{figure*}

Figure~\ref{fig_EvalFuncPR} shows a graph of the precision rate and recall rate for
 data sets with an acquisition rate of 0.1. 
However, since different data sets have different precision rates and recall rates,
 the precision rates and recall rates are scaled
 where the values of searches using L2 distance are 1.
The number of training data pairs used for training M-LSH was 2,000. 
Since Fig.~\ref{fig_EvalFuncPR} shows the scaled precision rate or the scaled recall rate,
 larger values indicate better performance from the learning method.


Although degraded in Fig.~\ref{fig_EvalFuncPR}, the M-LSH performance obtained using
 RATIO or COSINE\_RATIO is more or less unchanged from that of LSH.
Using COUNT, M-LSH performs better for all data sets.
Using COSINE, M-LSH performs worse for all data sets.


The reason why M-LSH using RATIO or COSINE\_RATIO has almost the same performance as LSH is
 thought to be as follows. 
The evaluation function includes a parameter $T$ that is analogous to temperature. 
Since we used a fixed value of $T=1$ in this evaluation,
 the index of the evaluation function is confined to the range $[0,2]$ or $[0,4]$. 
In this range, a slight change of the normal vector will not cause a large change
 in the value of the evaluation function. 
Therefore, the normal vector moves about more or less at random, so no large difference from LSH is obtained. 
In other words, in this evaluation function it can be said that $T=1$ corresponds to a high temperature. 
To increase the performance of the evaluation function,
 we should use a smaller $T$
 (i.e., a lower temperature), and expand the range of the evaluation function index to
 make the maximum value peaks sharper. 
However, at this limit, it can be approximated by COUNT. For this reason, at the low temperature limit,
 it is thought that these two evaluation functions  exhibit more or less the same performance
 as M-LSH when using COUNT.

\begin{figure*}[htb]
	\begin{center}
 	\includegraphics[scale=0.35]{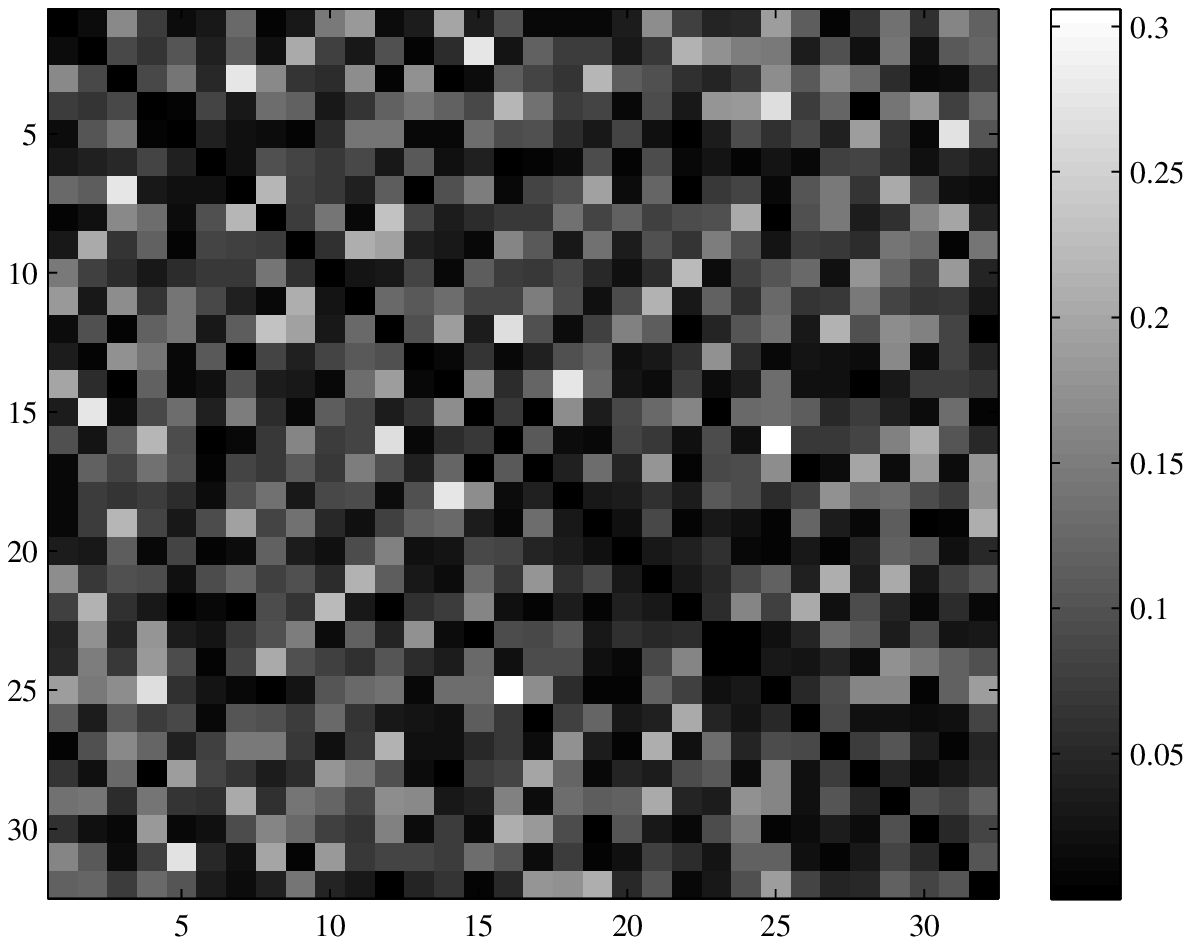}
 	\includegraphics[scale=0.35]{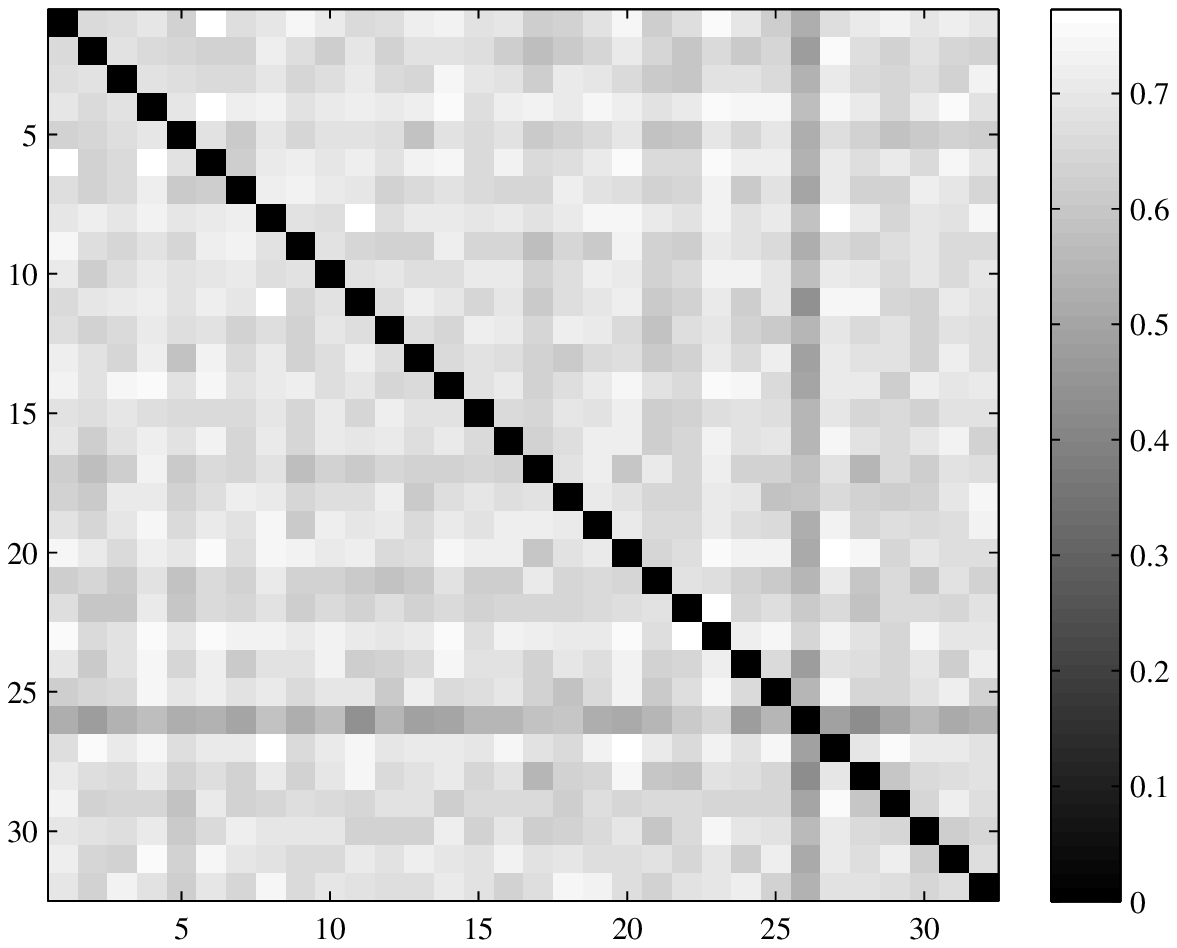}
 	\end{center}
	\caption{Cosines between normal vectors in M-LSH results learned for MNIST,
		with COUNT (left) and COSINE (right) used as the evaluation function.
	}
	\label{fig_CosineBetweenNormalVec}
\end{figure*}

It can be seen that COSINE performed much worse than LSH for the following reason. 
In COSINE, there is a gentle evaluation function gradient at all points in the region where the normal vector
 is defined. Therefore, the normal vectors tend to be oriented toward
 the point that shows a global maximum value. 
To see that the normal vector actually exists at a point showing the maximum value,
 we calculated the absolute value of the cosine between normal vectors. 
A larger absolute value of the cosine means that the vectors are pointing in similar directions. 
Figure~\ref{fig_CosineBetweenNormalVec} shows the absolute values of the cosines made
 by M-LSH normal vectors using 32-bit COUNT or COSINE values. 
In this figure, a matrix is calculated with the absolute values of cosines between 32-bit normal vectors
 as its constituent values, and these values are represented as a grayscale image. 
The diagonal elements are all zero. 
As Fig.~\ref{fig_CosineBetweenNormalVec} clearly shows, almost all of the normal vectors obtained
 with M-LSH are oriented in similar directions. 
In the case of COSINE, it is thought that the performance can be improved by taking a low temperature limit,
 as was the case for RATIO and COSINE\_RATIO. 
However, at this limit, COSINE can be approximated by COUNT. Furthermore, since COSINE requires
 more processing time than COUNT, there is no need to bother using COSINE.


Based on the above calculation results and discussion, it is thought that using COUNT
 as the evaluation function is more appropriate from the viewpoint of processing time and performance.

\subsection{Evaluation of sampling methods}
\label{subsec_resultSampling}


Here, we evaluate the performance of the sampling methods discussed in subsection~\ref{subsec_Sampling}. 
From the discussion of subsection~\ref{subsec_resultEvalFunc},
 we use the M-LSH method with the COUNT evaluation function to evaluate the performance of the sampling methods.

\begin{figure*}[htb]
	\begin{center}
 	\includegraphics[scale=0.35]{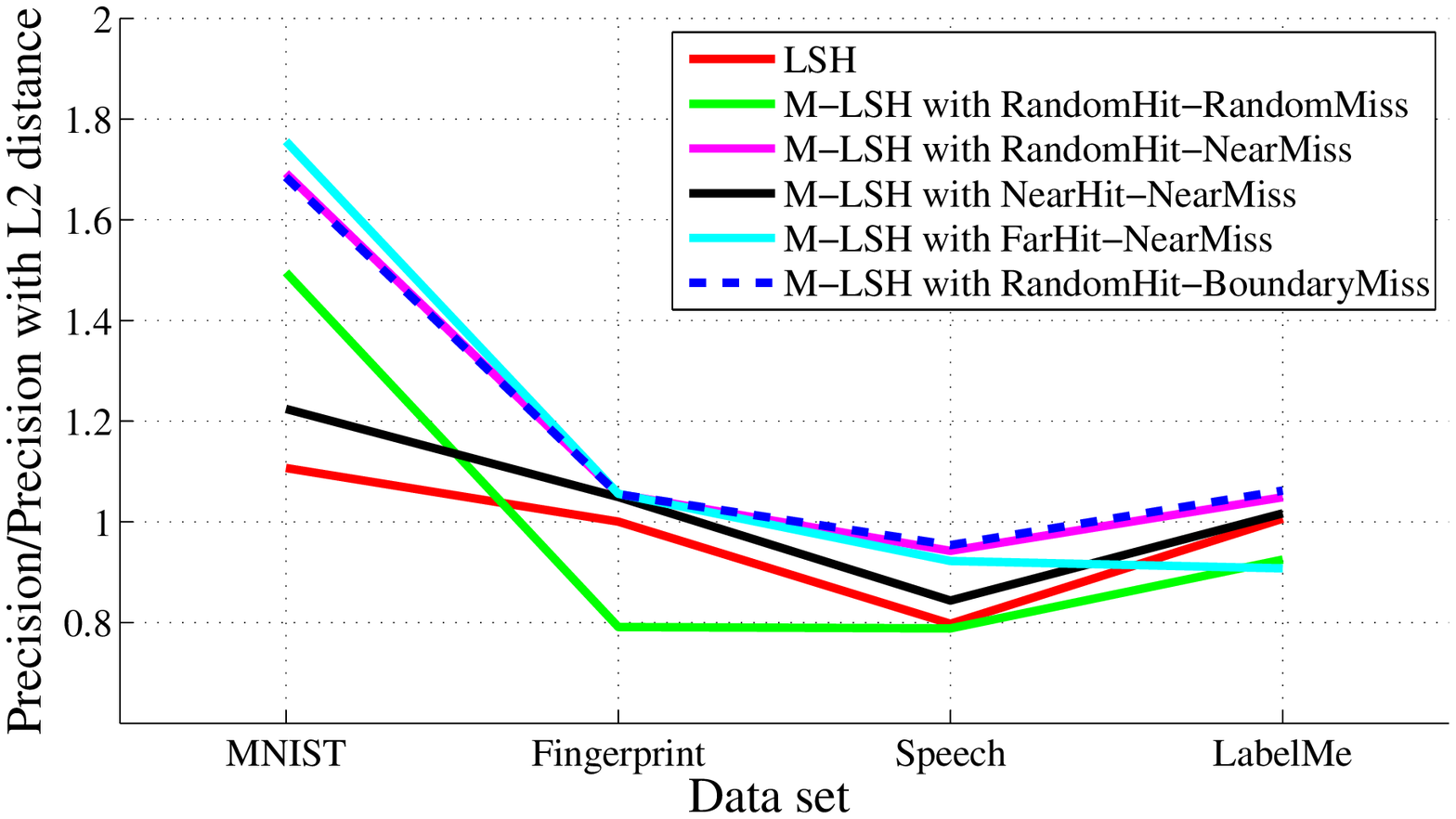}
 	\includegraphics[scale=0.35]{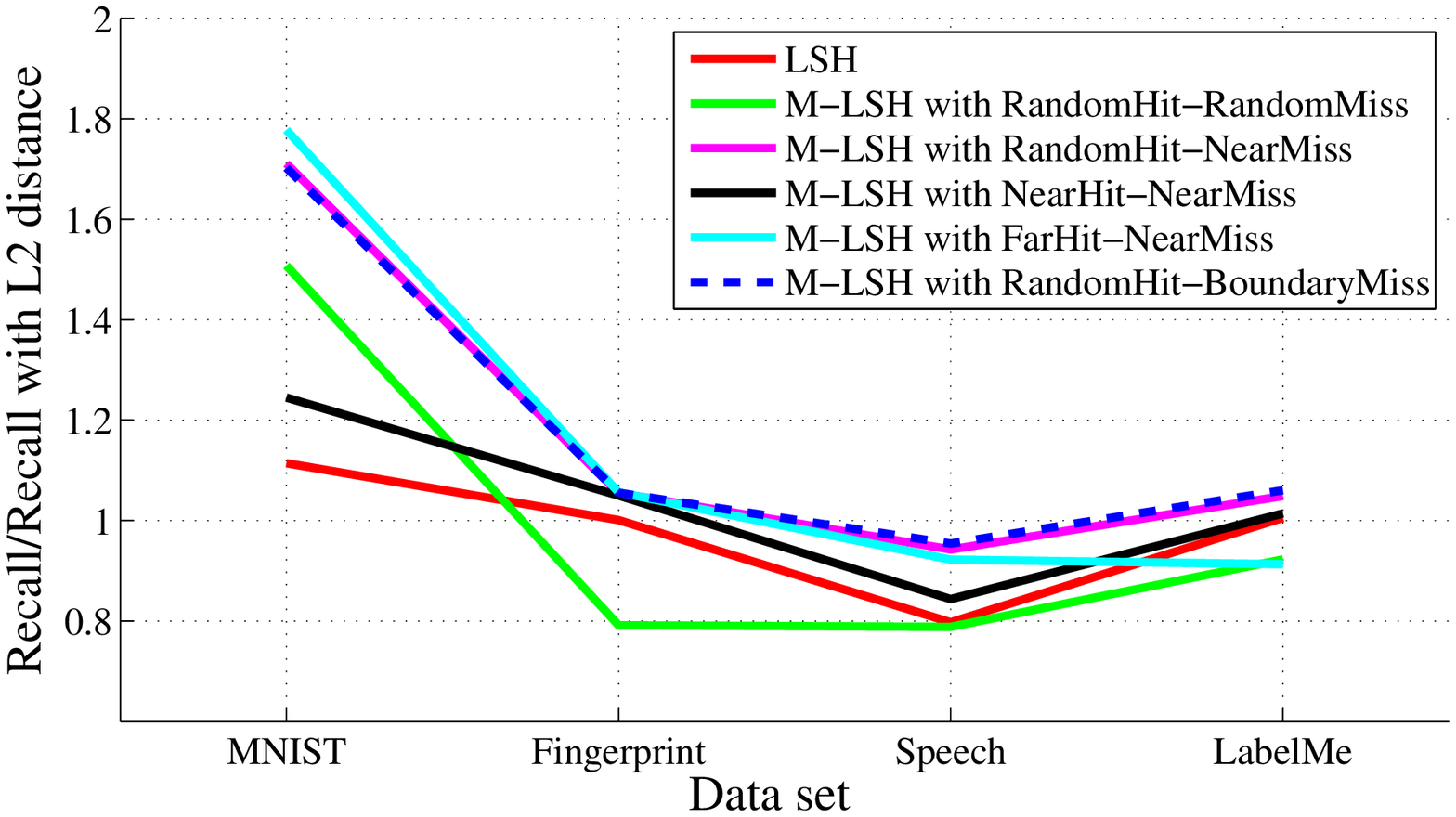}
 	\end{center}
	\caption{Performance of sampling methods with 2,000 items of training data:
		(Left) Data set and precision rate, (Right) Recall rate
	}
	\label{fig_SamplingPR}
	\begin{center}
 	\includegraphics[scale=0.35]{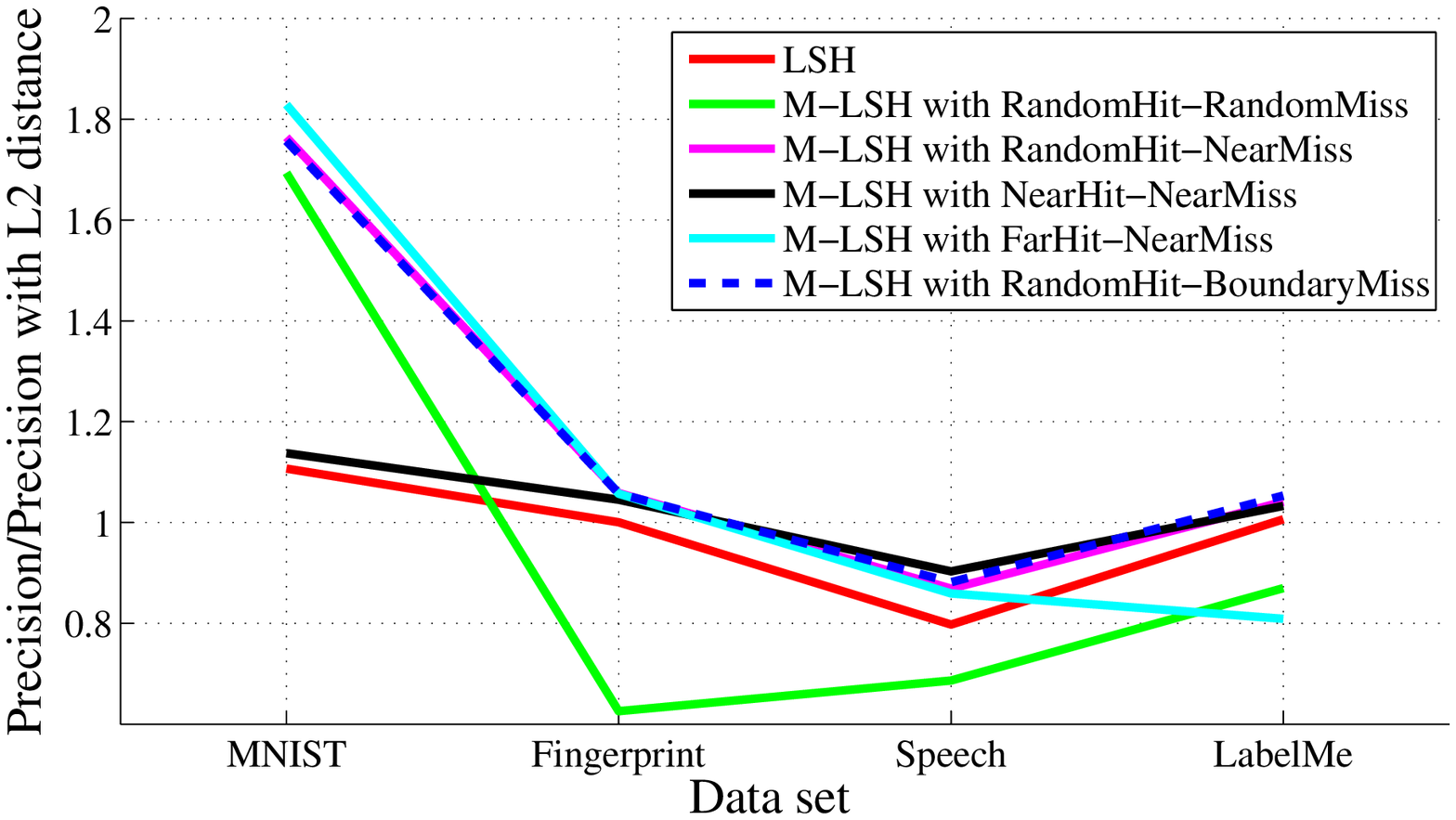}
 	\includegraphics[scale=0.35]{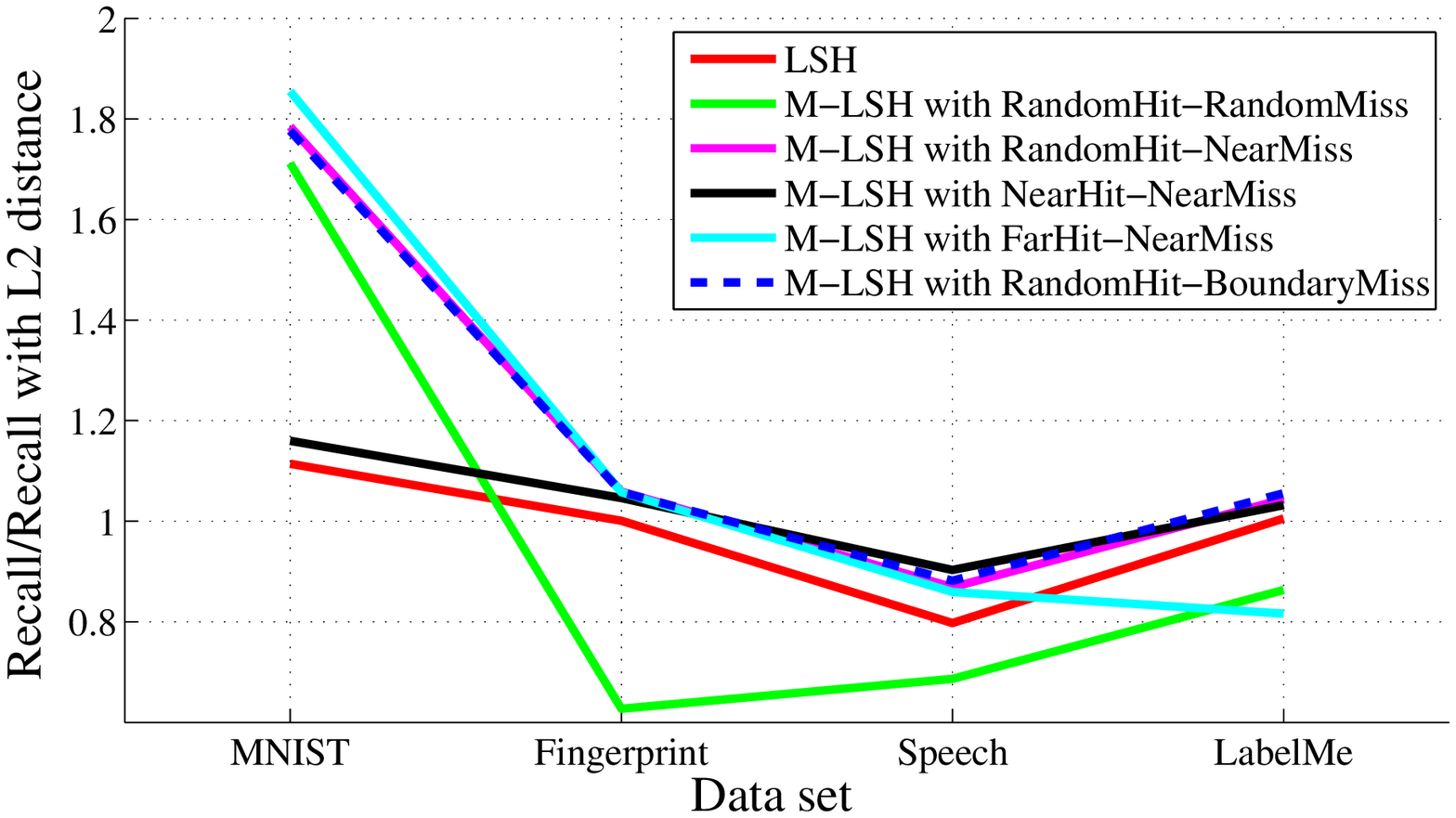}
 	\end{center}
	\caption{Performance evaluation of sampling methods with 20,000 items of training data:
		(Left) Data set and precision rate, (Right) Recall rate
	}
	\label{fig_SamplingSampleDaiPR}
\end{figure*}

In the same way as when evaluating the performance of the evaluation functions,
 we consider the scaled precision rate and scaled recall rate when the acquisition rate is 0.1. 
Figure~\ref{fig_SamplingPR} shows a graph of the scaled precision rate and recall rate of
 each sampling method in M-LSH using the COUNT evaluation function with 1,024 bits 
(except in the batch processing where the number of sample data items used was 2,000.)
To evaluate the dependence on the number of sample data items used for training,
 we also calculated the precision rate and recall rate with 20,000 sample data items,
 as shown in Fig.~\ref{fig_SamplingSampleDaiPR}.


From Figs.~\ref{fig_SamplingPR} and~\ref{fig_SamplingSampleDaiPR},
 it can be seen that the performance of M-LSH using Randomhit-Randommiss sampling is
 very poor for methods other than MNIST. 
The performance is worse than that of the LSH method.


The performance of M-LSH using Farhit-Nearmiss sampling was the best for MNIST. 
However, it had the worst performance for LabelMe.


It can be seen that the performance of M-LSH with the Nearhit-Nearmiss sampling method
 is depends strongly on the number of training data pairs. 
For the speech features and LabelMe data sets, the performance improves as the number
 of training data pairs increases. 
This performance improvement is thought to be due to the low probability of there being
 overlapping data pairs. 
For MNIST, the performance decreases as the number of training data pairs increases. 
This effect is thought to occur in the following way. 
As mentioned above, the role of positive pairs is to prevent data sets with a common label
 from being split by hyperplanes. 
It is therefore desirable that positive pairs are widely distributed across data sets having a common label. 
Nearhit sampling creates positive pairs by choosing the closest feature vectors with a common label,
 so a large number of positive pairs are needed for the distribution of a data set having a common label
 to be satisfied with a positive pair. 
In particular, MNIST requires more positive pairs than other data sets because there are
 a great many data items that have the same label.
The role of negative pairs is to separate data sets having different labels. 
Therefore, a number of negative pairs roughly equal to the square of the number
 of labels is sufficient. 
Since the positive pairs and negative pairs were used in equal numbers in these experiments,
 it seems that the effect of negative pairs in separating data sets having different labels
 outweighed the effect of positive pairs in preventing the separation of data sets having a common label. 
We think this is the reason why the performance decreases as the number of training data pairs is increased.


The M-LSH method using Randomhit-Nearmiss sampling and Randomhit-Boundarymiss sampling
 performed well for all data sets, regardless of the number of training data pairs. 
For MNIST and fingerprint images, the performance improves as the number of training data pairs is increased. 
However, for the speech features and LabelMe data sets,
 the performance was found to decrease as the number of training data pairs is increased. 
This is thought to be due to an increase in the number of overlapping data pairs. 
No large differences could be observed between these two sampling methods. 
However, for the speech features and LabelMe data sets, the performance was very slightly better
 with M-LSH using Randomhit-Boundarymiss sampling.


Based on these results, it is thought that the appropriate choice
 of sampling method depends on the properties of the data. 
Of the sampling methods we tried out in this study,
 it seems that the following choices are robust methods.
\begin{itemize}
\item
For data sets where each feature vector has unique label:
 Randomhit-Nearmiss sampling or Randomhit-Boundarymiss sampling
\item
For data sets where each feature vector has multiple labels and there are not many training data pairs:
 Randomhit-Boundarymiss sampling
\item
For data sets where each feature vector has multiple labels and there are very many training data pairs:
 Nearhit-Nearmiss sampling
\end{itemize}

\section{Comparison with existing learning methods}
\label{sec_kizontonosa}


In this section, we compare the performance of M-LSH with that of the existing learning methods LSH,
 MLH, and S-LSH. 
M-LSH uses the COUNT evaluation function and the Randomhit-Boundarymiss sampling method. 
The number of sample data pairs is 1,000 for both the positive pairs and negative pairs.

\begin{figure*}[htb]
	\begin{center}
 	\includegraphics[scale=0.35]{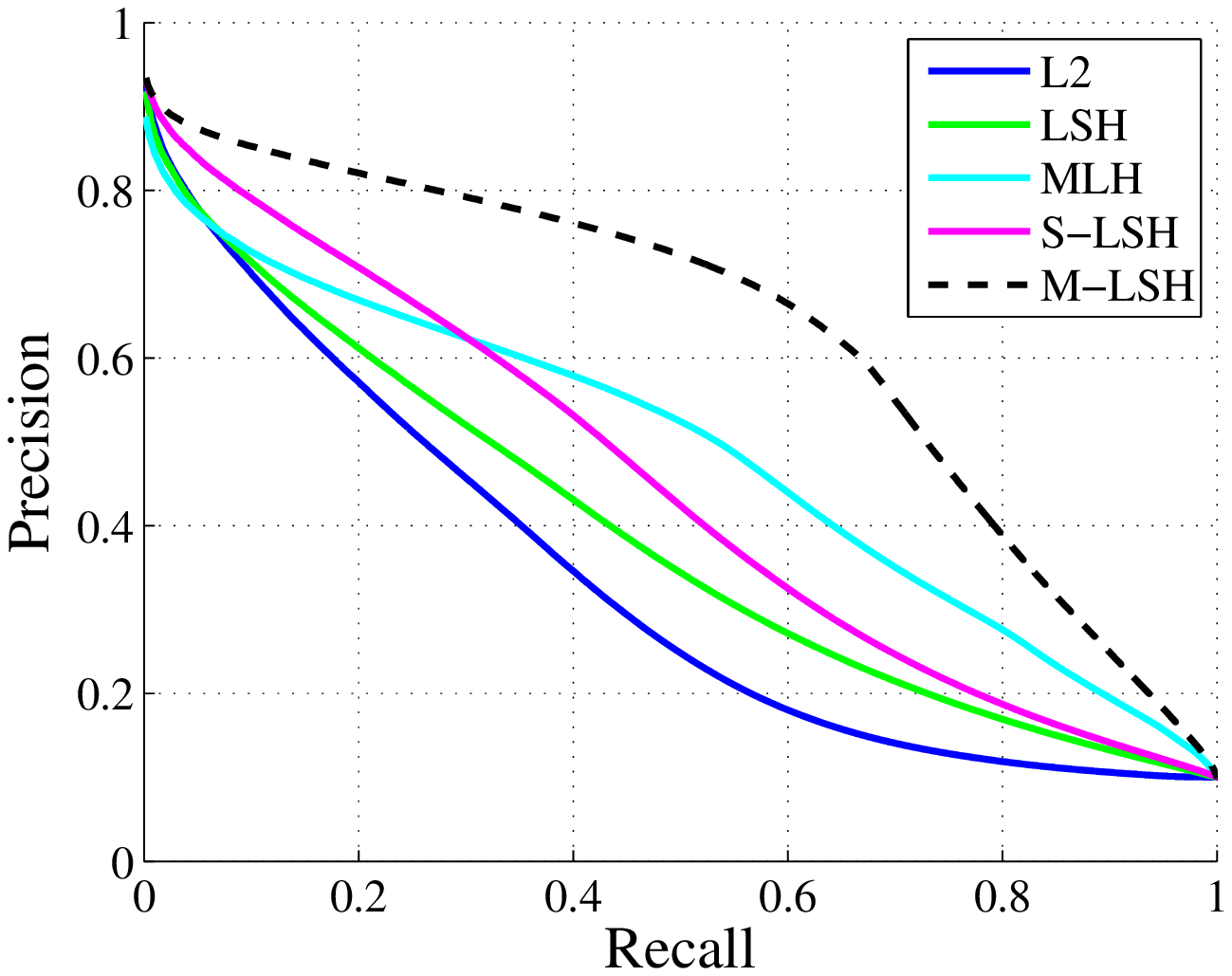}
 	\includegraphics[scale=0.35]{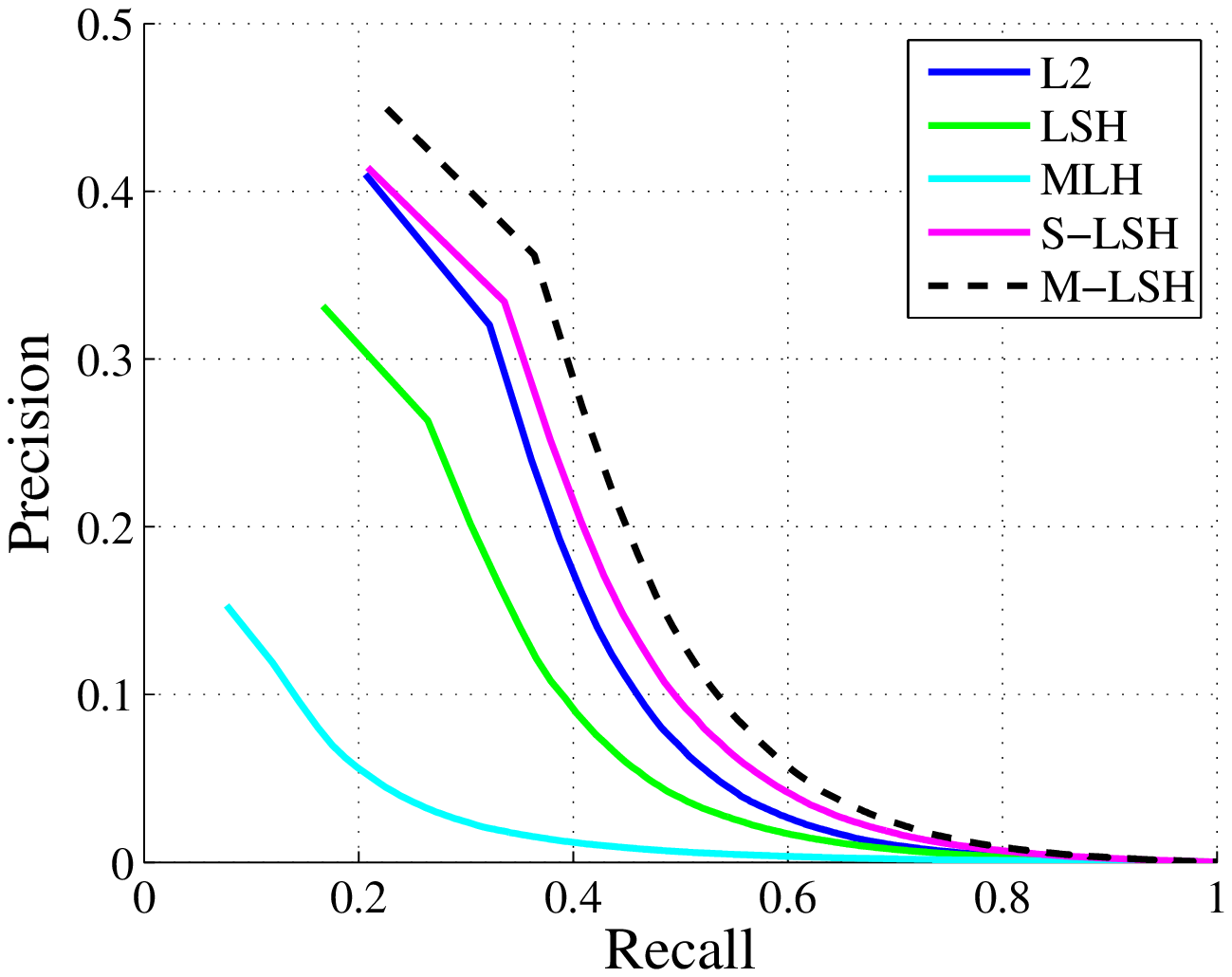}
 	\\
 	\includegraphics[scale=0.35]{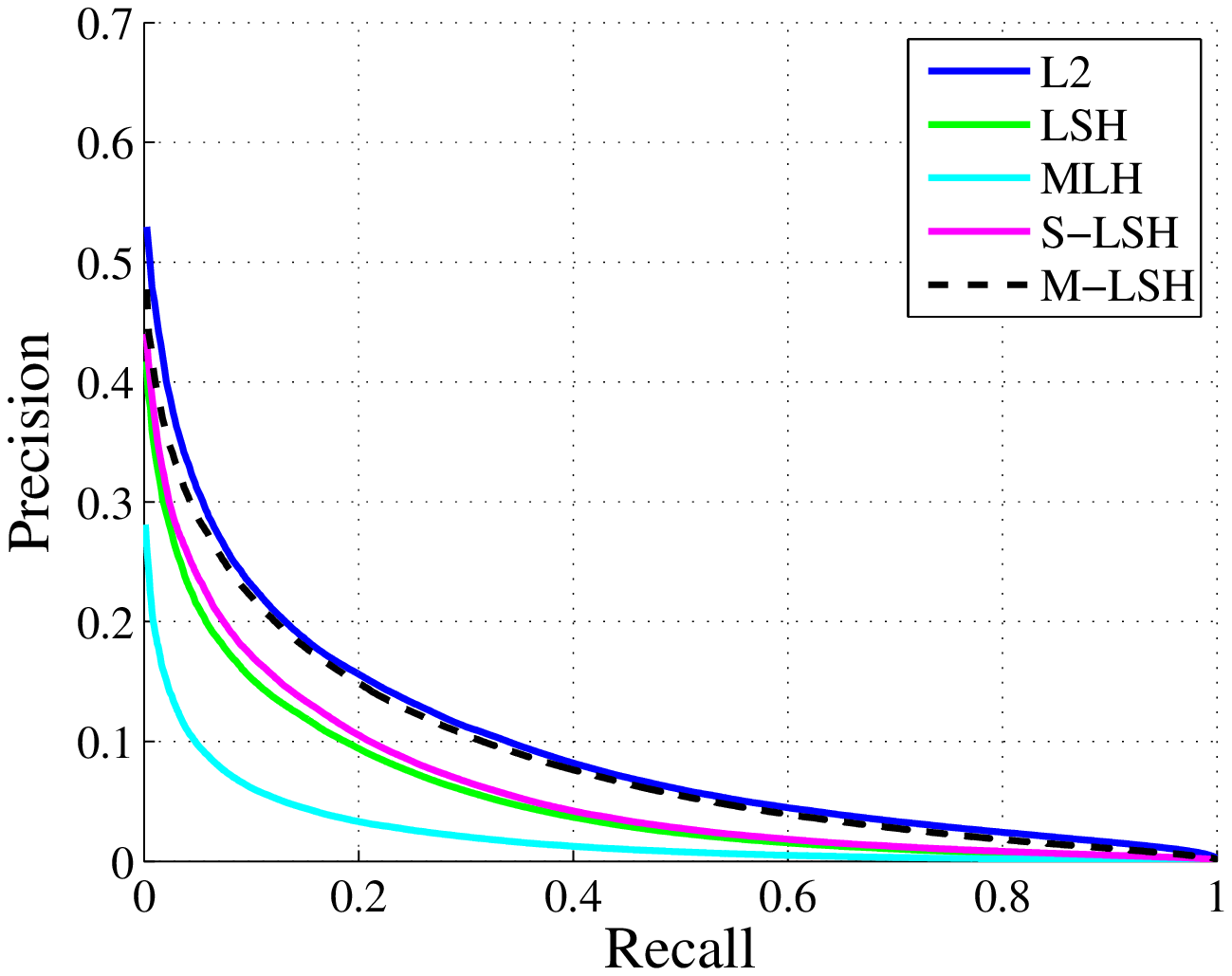}
 	\includegraphics[scale=0.35]{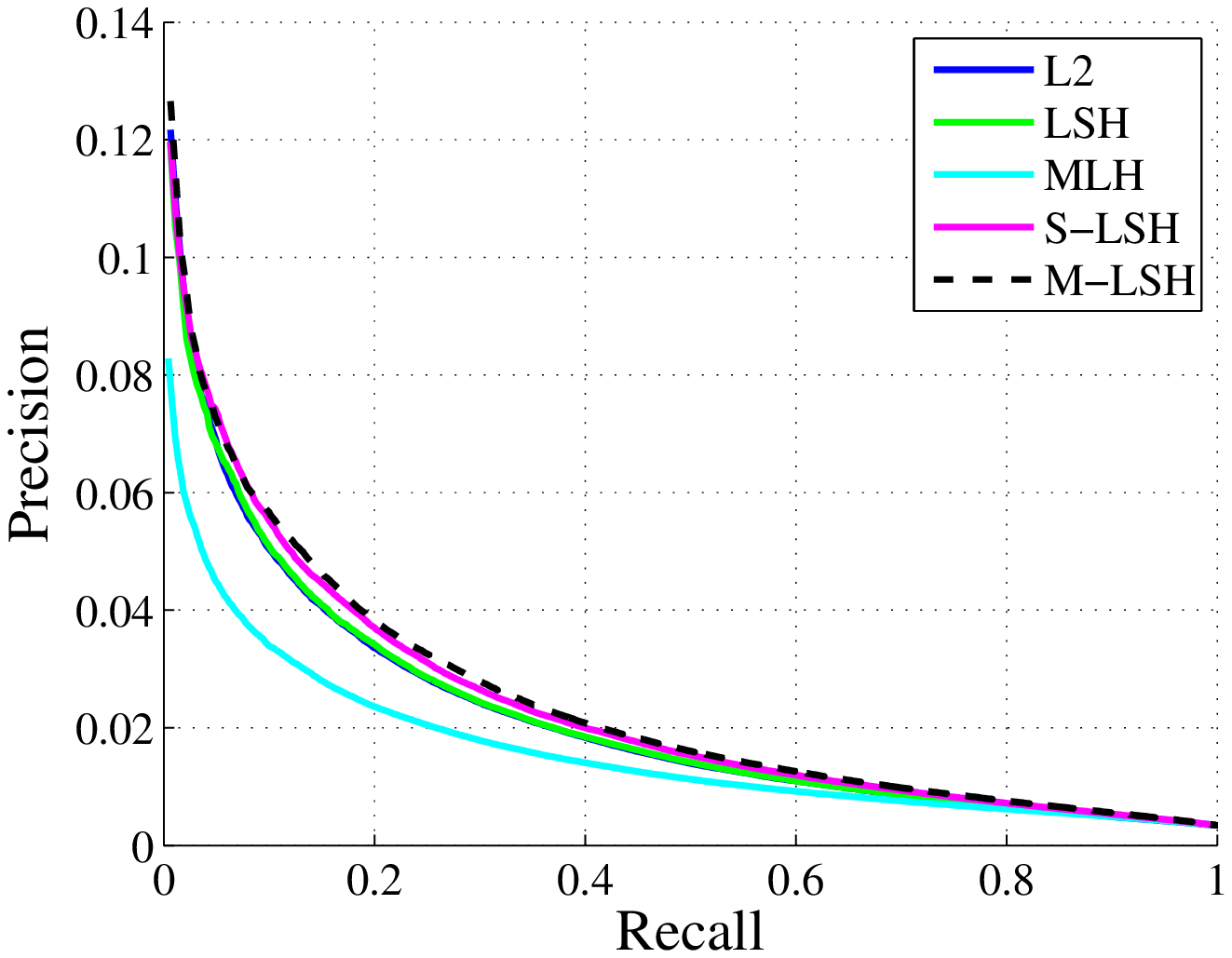}
 	\end{center}
	\caption{Recall-Precision curve for MNIST (upper left), fingerprint (upper right),
		speech (lower left), and LabelMe (lower right). 
	}
	\label{fig_PR}
\end{figure*}

Figure~\ref{fig_PR} shows the Recall-Precision curves for various different data sets. 
Here, the number of bits is 1,024. From these results,
 it can be seen that M-LSH outperforms the existing learning methods for all the data sets apart from LabelMe. 
In LabelMe, there are small regions where the S-LSH curve rises
 above the precision and recall curves for M-LSH,
 but it can be said that better overall performance is obtained with M-LSH.


From the above results, it is concluded that the proposed M-LSH learning method has good hashing performance.

\section{Summary and future works}
\label{sec_conclusion}


In this paper, we proposed a learning method for hyperplanes using MCMC. 
We also considered evaluation functions and sampling methods used in this learning method,
 and we evaluated their performance. 
As a result, we have confirmed that this proposed method exceeds the performance of existing learning methods.


Finally, we mention the direction of future research. 
When using the MCMC method for learning, the ultimate positions of particles do not lie at points
 that maximize the evaluation function. 
One way in which this problem could be resolved involves recording the particle loci and finding out
 which point maximizes the evaluation function.

\bibliographystyle{unsrt}
\bibliography{BibForHashing}

\begin{thebibliography}{10}

\bibitem{KDTree}
Sunil Arya, David~M. Mount, Nathan~S. Netanyahu, Ruth Silverman, and Angela~Y.
  Wu.
\newblock An optimal algorithm for approximate nearest neighbor searching fixed
  dimensions.
\newblock {\em J. ACM}, 45(6):891--923, nov 1998.

\bibitem{iDistance}
H.~V. Jagadish, Beng~Chin Ooi, Kian-Lee Tan, Cui Yu, and Rui Zhang.
\newblock idistance: An adaptive b+-tree based indexing method for nearest
  neighbor search.
\newblock {\em ACM Trans. Database Syst.}, 30(2):364--397, jun 2005.

\bibitem{Weber1998DimensionCurse}
Roger Weber, Hans-J\"{o}rg Schek, and Stephen Blott.
\newblock A quantitative analysis and performance study for similarity-search
  methods in high-dimensional spaces.
\newblock In {\em Proceedings of the 24rd International Conference on Very
  Large Data Bases}, VLDB '98, pages 194--205, San Francisco, CA, USA, 1998.
  Morgan Kaufmann Publishers Inc.

\bibitem{LSH_IndykMotwani}
Piotr Indyk and Rajeev Motwani.
\newblock Approximate nearest neighbors: towards removing the curse of
  dimensionality.
\newblock In {\em Proceedings of the thirtieth annual ACM symposium on Theory
  of computing}, STOC '98, pages 604--613, New York, NY, USA, 1998. ACM.

\bibitem{LSH_RandomProjection}
Moses~S. Charikar.
\newblock Similarity estimation techniques from rounding algorithms.
\newblock In {\em Proceedings of the thiry-fourth annual ACM symposium on
  Theory of computing}, STOC '02, pages 380--388, New York, NY, USA, 2002. ACM.

\bibitem{LSH_FS1}
Yadong Mu, Xiangyu Chen, Tat-Seng Chua, and Shuicheng Yan.
\newblock Learning reconfigurable hashing for diverse semantics.
\newblock In {\em Proceedings of the 1st ACM International Conference on
  Multimedia Retrieval}, ICMR '11, pages 7:1--7:8, New York, NY, USA, 2011.
  ACM.

\bibitem{LSH_FS2}
Yu-Gang Jiang, Jun Wang, and Shih-Fu Chang.
\newblock Lost in binarization: query-adaptive ranking for similar image search
  with compact codes.
\newblock In {\em Proceedings of the 1st ACM International Conference on
  Multimedia Retrieval}, ICMR '11, pages 16:1--16:8, New York, NY, USA, 2011.
  ACM.

\bibitem{PCAH}
Xin-Jing Wang, Lei Zhang, Feng Jing, and Wei-Ying Ma.
\newblock Annosearch: Image auto-annotation by search.
\newblock In {\em Proceedings of the 2006 IEEE Computer Society Conference on
  Computer Vision and Pattern Recognition - Volume 2}, CVPR '06, pages
  1483--1490, Washington, DC, USA, 2006. IEEE Computer Society.

\bibitem{MLH}
Mohammad Norouzi and David~J. Fleet.
\newblock Minimal loss hashing for compact binary codes.
\newblock In Lise Getoor and Tobias Scheffer, editors, {\em ICML}, pages
  353--360. Omnipress, 2011.

\bibitem{SIMBA_KN}
Makiko Konoshima and Yui Noma.
\newblock Locality-sensitive hashing with margin based feature selection.
\newblock {\em arXiv:1209.5833}, 2012.

\bibitem{LIFT_KN}
Yui Noma and Makiko Konoshima.
\newblock Hyperplane arrangements and locality-sensitive hashing with lift.
\newblock {\em arXiv:1212.6110}, 2012.

\bibitem{Hastings_1970}
W~K Hastings.
\newblock Monte carlo sampling methods using markov chains and their
  applications.
\newblock {\em Biometrika}, 57(1):97--109, 1970.

\bibitem{MNIST}
The {MNIST} database of handwritten digits.
  \url{http://yann.lecun.com/exdb/mnist/}.

\bibitem{KawasakiCouncil}
Internet relay broadcast of {K}awasaki-city parliament.
  \url{http://www.kawasaki-council.jp/}.

\bibitem{GIST}
Aude Oliva and Antonio Torralba.
\newblock Modeling the shape of the scene: A holistic representation of the
  spatial envelope.
\newblock {\em Int. J. Comput. Vision}, 42(3):145--175, may 2001.

\bibitem{512dimGISTdata}
Antonio Torralba, Rob Fergus, and Yair Weiss.
\newblock Small codes and large image databases for recognition.
\newblock In {\em In Proceedings of the IEEE Conf on Computer Vision and
  Pattern Recognition}, 2008.

\end{thebibliography}


\end{document}